\documentclass[journal]{IEEEtran}
\usepackage{cite}
\usepackage{hyperref}

\ifCLASSINFOpdf
\usepackage[pdftex]{graphicx}
\else
\fi
\usepackage{amsfonts,amsmath,amssymb,amsthm}
\usepackage{graphicx}
\usepackage{subcaption}
\usepackage{caption}
\usepackage{color}

\begin{document}
%
\title{Pattern Integration and Enhancement Vision Transformer for Self-supervised Learning in Remote Sensing}
%
%
%

\author{Kaixuan~Lu,
Ruiqian~Zhang,
        Xiao~Huang,
        Yuxing~Xie,
        Xiaogang~Ning,
        Hanchao~Zhang,
        Mengke~Yuan,
        Pan~Zhang,
        Tao~Wang,
        Tongkui~Liao
\thanks{The work of Ruiqian Zhang is supported by the National Natural Science Foundation of China (grant number: 42201440) and the Fundamental Research Funds for Chinese Academy of Surveying and Mapping (grant numbers: AR2201 and AR2410). The work of Yuxing Xie is supported by Shenzhen Science and Technology Program (grant number: ZDSYS20210929115800001).}
\thanks{Kaixuan Lu, Mengke Yuan, Pan Zhang, Tao Wang and Tongkui Liao are with Piesat Information Technology Co., Lrd., Beijing 100195, China(e-mail:lukaixuan, yuanmengke, zhangpan, wangtao, liaotongkui@piesat.cn).}
\thanks{Ruiqian Zhang, Xiaogang Ning and Hanchao Zhang are with the Chinese Academy of Surveying and Mapping, 100036 Beijing, China (e-mail: zhangrq, ningxg, zhanghc@casm.ac.cn).}
\thanks{Xiao Huang is with the Department of Environmental Sciences, Emory University, Atlanta, GA 30322, USA (xiao.huang2@emory.edu).}
\thanks{Yuxing Xie is with Research Institute of Urbanization and Urban Safety, School of Civil and Resource Engineering, University of Science and Technology Beijing, 100083 Beijing, China (e-mail: yuxing.xie@outlook.com).}}

\maketitle

\begin{abstract}
Recent self-supervised learning (SSL) methods have demonstrated impressive results in learning visual representations from unlabeled remote sensing images. However, most remote sensing images predominantly consist of scenographic scenes containing multiple ground objects without explicit foreground targets, which limits the performance of existing SSL methods that focus on foreground targets. This raises the question: Is there a method that can automatically aggregate similar objects within scenographic remote sensing images, thereby enabling models to differentiate knowledge embedded in various geospatial patterns for improved feature representation? In this work, we present the Pattern Integration and Enhancement Vision Transformer (PIEViT), a novel self-supervised learning framework designed specifically for remote sensing imagery. PIEViT utilizes a teacher-student architecture to address both image-level and patch-level tasks. It employs a proposed, Geospatial Pattern Cohesion (GPC) module to explore the natural clustering of patches, enhancing the differentiation of individual features. A Feature Integration Projection (FIP) module is employed to further refine masked token reconstruction using geospatially clustered patches. We validated PIEViT across multiple downstream tasks, including object detection, semantic segmentation, and change detection. Experiments demonstrated that PIEViT enhances the representation of internal patch features, providing significant improvements over existing self-supervised baselines. It achieves excellent results in object detection, land cover classification, and change detection, underscoring its robustness, generalization, and transferability for remote sensing image interpretation tasks.
\end{abstract}


%
\IEEEpeerreviewmaketitle

\section{Introduction}
%
%
%
%
\IEEEPARstart{R}{emote} sensing technology, employing aerial or satellite platforms, facilitates the acquisition of earth observation data from afar, marking its utility across diverse sectors such as agriculture monitoring, environmental surveillance evaluation, urban planning, and disaster management. At the heart of these applications lies the interpretation of remote sensing imagery, a task that has been considerably challenging \cite{lillesand2015remote}. The advent of deep learning technology \cite{vaswani2017attention} has significantly enhanced the precision and automation of remote sensing image analysis, encompassing object detection, land cover classification, and change detection. Nonetheless, the intrinsic complexity of remote sensing scenarios—attributable to variations in sensor technology, atmospheric conditions, and image resolution—results in the performance disparity of analytical models across different contexts. The prohibitive costs associated with manual annotation further exacerbate this issue, rendering the re-labeling of samples for specific scenarios impractical. Consequently, there is a growing demand for models endowed with superior generalization capabilities, capable of adeptly navigating the multifaceted landscape of remote sensing applications.

In Self-Supervised Learning (SSL), algorithms learn useful features by generating labels or targets from the input data itself, facilitating an understanding of the data. A key advantage of this approach is its independence from manually annotated data, enabling more efficient utilization of large volumes of unlabeled data. In the realm of remote sensing, employing SSL for pre-training on a wide array of scene images enhances feature representation capabilities and generalizability across different scenarios. Through transfer learning, outstanding results can be achieved across various imagery-based, downstream tasks. Presently, contrastive learning \cite{wu2018unsupervised,ye2019unsupervised} and masked learning \cite{bao2021beit,he2022masked} stand as two widely employed SSL methods in both computer vision and remote sensing fields. In contrastive learning, models are trained to distinguish between a candidate sample and positive (similar) and negative (dissimilar) samples; in masked learning, the objective is to reconstruct obscured portions based on contextual information. Contrastive learning focuses on learning representations that differentiate between samples, while masked learning emphasizes the reconstruction or prediction of missing data based on partial information. Both methods have demonstrated significant potential in remote sensing applications. However, unlike natural images, remote sensing images are predominantly scenographic, often featuring multiple ground objects without a clear foreground target. This characteristic imposes certain limitations on the methods above. For instance, in contrastive learning, random cropping to generate positive sample pairs may result in two images with completely dissimilar features due to the scenographic nature of remote sensing images, potentially leading to semantic confusion in the model. Pixel-level masked learning, on the other hand, entirely disregards semantic information.

Seeking a method that discerns both the semantic information of different scene images and the semantic details within each area of the images is essential for effectively processing scenographic remote sensing imagery. It has been observed that scenographic images in remote sensing typically exhibit specific distribution patterns, where natural elements (such as mountains, bodies of water, forests, lakes, and grasslands) and man-made elements (such as buildings, farmlands, plowed lands, and gardens) often display pronounced clustering. This tendency for feature aggregation leads to the formation of discernible geographic pattern clusters among individual image patches and their neighbors, thereby establishing intricate spatial relationships across the varying patches and their adjacent areas.

Building on these observations, we introduce a novel SSL feature learning algorithm named the Pattern Integration and Enhancement Vision Transformer. PIEViT consists of a student network and a teacher network, simultaneously addressing image-level and patch-level tasks. For the image-level task, we compute the cross-entropy loss between the class tokens extracted by the student and teacher networks from randomly cropped versions of the same image. For the patch-level task, we randomly mask some patch input to the student network and then calculate the cross-entropy loss between each masked patch token and the corresponding patch token in the teacher network. By integrating image-level loss with patch-level loss and backpropagating through the student network's parameters, the teacher network is updated using the Exponential Moving Average (EMA) strategy \cite{caron2021emerging}. To better differentiate the features of each patch within images, we propose the Geospatial Pattern Cohesion (GPC) module to explore and leverage the natural clustering of landscape elements observed in remote sensing imagery, substantially augmenting the model's proficiency in decoding intricate spatial distributions. Furthermore, we present the Feature Integration Projection (FIP) module, which utilizes the Geospatial Pattern Cohesion scores generated by the teacher network to guide and refine the reconstruction of masked tokens in the student network. This approach not only increases the complexity of the learning task but also bolsters the learning of nuanced differences in internal features across the images.
Overall, the primary contributions of this work can be summarized as follows:
\begin{itemize}
    \item We introduce the Pattern Integration and Enhancement Vision Transformer (PIEViT), which supervises the generation of masked patches in the student branch with features aggregated from geospatially clustered patches in the teacher branch. PIEViT enhances the distinctiveness of internal area features within images, achieving superior performance in downstream tasks.

    \item Unlike the simple Multilayer Perceptron (MLP)-based projection head, our method computes the score from the Geospatial Pattern Cohesion (GPC) module between each Query Patch and its neighboring patches, selects the top $k$ patches based on their scores, and then processes these patches through a Feature Integration Projection (FIP) module. Compared to MLP, FIP is more adept at capturing a broader range of patch-level semantic information.

    \item After pretraining on the unlabeled dataset, we validated our approach across multiple downstream tasks. PIEViT possesses a strong capability to represent rich semantic and local information, achieving SOTA results in object detection, land cover classification, and change detection, in comparison with classical models with similar size. This demonstrates its generalization and transferability for remote sensing image interpretation tasks.

\end{itemize}

\section{Related Works}
\subsection{Contrastive Learning}
Self-supervised contrastive learning methods train a model by contrasting pairs of inputs derived from the same image. By design, these methods usually employ Siamese architectures. In detail, they usually utilize data augmentation to generate various views of the same image. These augmented views are then fed into different network branches to obtain corresponding representations. By training the model to differentiate between these representations, contrastive learning approaches can facilitate network training and improve the model’s ability to capture higher-level features that can be generalized to downstream tasks \cite{he2020momentum,chen2020simple,chen2020improved,chen2021empirical}. Most contrastive methods originate from computer vision tasks that concentrate on small areas. Therefore, they are primarily developed based on image-level objectives \cite{chen2021exploring,caron2020unsupervised,grill2020bootstrap,zhou2021ibot}. However, for remote sensing imagery, in addition to image-level information, there is a critical need for local patch-level information, due to the reason that remote sensing images cover large-area (e.g., city-scale) regions. 

Recent works in contrastive learning have focused on the excavation of local feature information. For instance, dense contrastive learning (DenseCL) \cite{wang2021dense} achieves self-supervised learning by optimizing a pairwise contrastive (dis)similarity loss at the pixel level between two views of an input image. Selecting suitable negative samples at the pixel level presents a challenge for DenseCL due to the limited information contained in individual pixels. Object-level representation learning (ORL) \cite{xie2021unsupervised} applies unsupervised region proposals and kNN-based object similarity to diversify both intra-image and inter-image with more variance, aiding in object-level representation learning. However, it requires three stages of pre-training, leading to less stable model performance. DetCo \cite{xie2021detco} aims to enhance the performance of general contrastive learning methods by augmenting multiple global and local views simultaneously, yet it is primarily optimized for object detection tasks. For non-detection tasks (e.g., segmentation), it may not be as effective as other contrastive learning methods. Region similarity representation learning (ReSim) \cite{xiao2021region} leverages contrastive learning to obtain the similarity of different feature regions from varying perspectives through RoI Pooling, but the model's performance is highly sensitive to the selection of regions. If the chosen regions are not representative or insufficient to capture the image's key features, it could adversely affect the learning outcomes. Considering that a set of pixel-wise features contains more semantic and structure information, set similarity (SetSim) \cite{wang2022exploring} generalizes pixel-wise similarity learning to set-wise similarity. This method effectively improves robustness through set similarity between views, using attention features to establish corresponding sets, filtering out noise backgrounds, and addressing misleading pixel-level features and semantic inconsistencies. However, pixel-level correspondences based on similarity or geometric information are prone to noise, a problem that is pronounced in remote sensing images.

In short, while these methods represent advancements in the extraction of pixel-level and region-level features, they encounter specific challenges when applied to remote sensing imagery, such as selecting effective negative samples, achieving stable model performance through multi-stage training, optimizing for non-detection tasks, sensitivity to region selection, and handling noise in pixel-level correspondences. This underscores the complexity of addressing spectral anomalies and spectral homogeneity across different objects in remote sensing images.

\subsection{Masked Image Modeling (MIM)}
Inspired by the BERT \cite{devlin2018bert} from the natural language processing domain, the MIM approach learns visual representations through the reconstruction of masked patches. Based on the reconstruction target, it can be divided into patch-level, e.g., BeiT \cite{bao2021beit} and pixel-level, e.g., MAE \cite{he2022masked} approaches, where the former predicts the semantic tokens of masked patches, and the latter predicts the RGB values of masked patches. BeiT is the first self-supervised method in the visual domain to use patch-level MIM. It first tokenizes the image based on dVAE \cite{ramesh2021zero}, then obtains hidden layer features after passing the masked patches through an encoder, and finally completes token reconstruction by calculating the loss with softmax against the tokenized values. MAE \cite{he2022masked} and SimMIM \cite{xie2022simmim} are the first two self-supervised algorithms in the visual domain to use pixel-level MIM. MAE \cite{he2022masked} employs an encoder-decoder architecture, where the encoder processes unmasked image blocks, and the decoder reconstructs the entire image. SimMIM \cite{xie2022simmim} uses a more diversified architectural design, utilizing just an encoder or combining an encoder-decoder, and finally reconstructs the original pixel values of masked patches using a lightweight prediction head. In addition, CAE \cite{chen2024context} introduces a cross-attention module to learn latent space knowledge between visible and masked patches, while MaskFeat \cite{wei2022masked} proposes using the hand-crafted HOG features of masked patches as the prediction target in self-supervised pertaining.

The above self-supervised methods based on patch-level MIM primarily focus on the semantic recovery of local features but lack the acquisition of global semantic information. Pixel-level MIM methods focus on learning high-frequency spatial information, lacking in low-frequency semantic information, which leads to significant semantic confusion when transitioning to downstream tasks. 

\subsection{Self-supervised Learning in Remote Sensing Domain}
The application of SSL techniques is increasingly widespread in the remote sensing field, aiming to leverage unlabeled satellite imagery to learn robust and informative feature representations. These methods can also be broadly classified into two main categories: contrastive learning-based and MIM-based methods, similar to those in computer vision. Studies like \cite{jung2021contrastive,manas2021seasonal,noman2024rethinking} have applied contrastive SSL techniques to satellite imagery feature representation learning. Although these methods have achieved commendable results, they focus on temporal and spatial global self-supervised learning, neglecting the local semantic information of remote sensing images.

The MIM-based approaches are also extensive in the remote sensing domain. For instance, RingMo \cite{sun2022ringmo} proposes the PIMask Strategy based on MAE \cite{he2022masked}, which retains some pixels in the masked parts to better recover small objects. RVSA \cite{wang2022advancing}, leveraging MAE \cite{he2022masked} for self-supervised training, introduces rotated varied-size attention based on ViT \cite{dosovitskiy2020image} during the fine-tuning stage for downstream tasks to adapt to multi-directional remote sensing images. GFM \cite{mendieta2023building} builds on SimMIM \cite{xie2022simmim} by using a frozen ImageNet22k \cite{russakovsky2015imagenet} pre-trained model as a teacher network to obtain image features, which are then compared for similarity with features in the student network. SatMAE \cite{cong2022satmae} extends MAE technology to temporal and multi-spectral satellite imagery, with an innovative approach of using extra tokens to maintain temporal consistency. Building on this, SatMAE++ \cite{huang2024generic} further advances the pre-training of multispectral and RGB images at multiple scales. Scale-MAE \cite{reed2023scale} introduces the concept of scale into MAE, adhering to the principle that more refined ground objects require more pixels for description, by using scale-variant positional encoding, which scales the original positional encoding by 1/GSD. In addition, some methods, such as SpectralGPT \cite{hong2024spectralgpt} and S2HM2 \cite{tu2024s}, focus on SSL for hyperspectral images and design networks that cater to the unique spatial-spectral structure. These MIM methods proficiently capture the spatial texture information of remote sensing image features but lack semantic information, posing challenges for downstream remote sensing tasks. To address these limitations, we introduced PIEViT, which employs a teacher-student architecture and uses neighborhood similarity aggregation of patch features to predict masked patch tokens, while also retaining the ability to learn image-level global semantic information.

\section{Methodology}
\subsection{Approach Overview} \label{SEC:III-A}
\begin{figure*}[htpb]
\centering
\includegraphics[width=\textwidth]{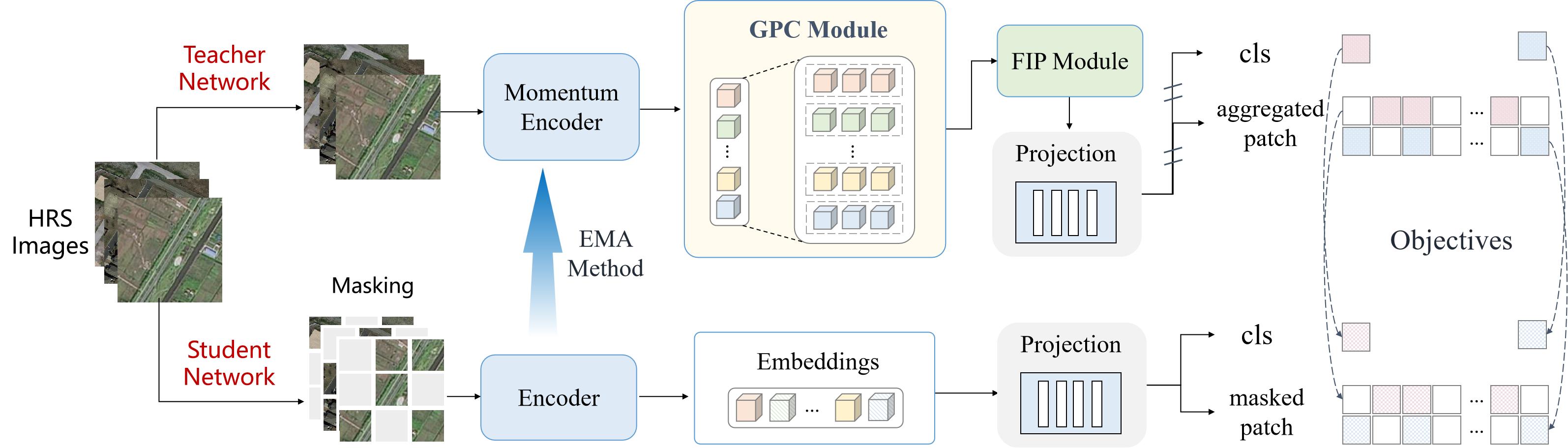}
\caption{PIE Network Structure, performing masked image modeling by self-distillation. Given two different views of an image, which are input into the teacher network and the student network respectively, we apply a stop-gradient (SG) operator in the teacher network, propagating gradients only through the student. The teacher parameters are updated using the exponential moving average (EMA) of the student parameters.}
\label{fig_1}
\end{figure*}

To facilitate spatiotemporal semantic learning in remote sensing images, we introduce PIEViT, a visual transformer designed for patch geospatial pattern representation learning that is specially crafted for the complexities of remote sensing data. PIEViT incorporates a dual-stream encoder as its backbone, exploiting the comprehensive feature extraction process of Vision Transformers (ViT) alongside self-distillation \cite{caron2021emerging} methodologies to enhance self-supervised learning. Moreover, we introduce two innovative modules: the Geospatial Pattern Cohesion (GPC) Module and the Feature Integration Projection (FIP). The former is engineered to enhance the learning of representations for internal scene components within remote sensing images, while the latter employs the inherent clustering pattern to supervise and refine the masked tokens in the student network, thus deepening the discrimination of complex spatial distributions. Consequently, PIEViT not only intensifies the learning task's complexity but also focuses on internal feature variations more acutely, resulting in a pre-trained model optimally configured for downstream tasks in remote sensing. The architecture of PIEViT is delineated in Figure \ref{fig_1}.

The architecture encompasses both a student and a teacher network, establishing a dual-stream feature learning framework. Despite sharing the same structural components, including an encoder and a projection head, each network operates with unique parameters. The teacher network's role is to set learning objectives for the student network. To this end, a stop-gradient (SG) mechanism is applied to the teacher network, inhibiting any gradient backpropagation through it and ensuring that only the student network is subject to backpropagation. The teacher network's parameters are dynamically updated through an EMA derived from the student network's parameters. In pursuit of greater input patch diversity, initial cropping is applied to the input images (sized $H\times W\times3$). Subsequently, for a specific image $x$, two randomly augmented versions, $u$ and $v$, are created, with selective patch masking yielding their masked counterparts, $u_m$ and $v_m$. These masked versions are then input into the student network, while their unmasked originals are directed to the teacher network. After being processed through the ViT encoder, a sequence of query patch embeddings is generated, representing the feature representation of each query patch, which encapsulates the rich semantic information extracted from the remote sensing imagery.

\subsection{The Geospatial Pattern Cohesion (GPC)}
\begin{figure*}[htpb]
\centering
\includegraphics[width=\textwidth]{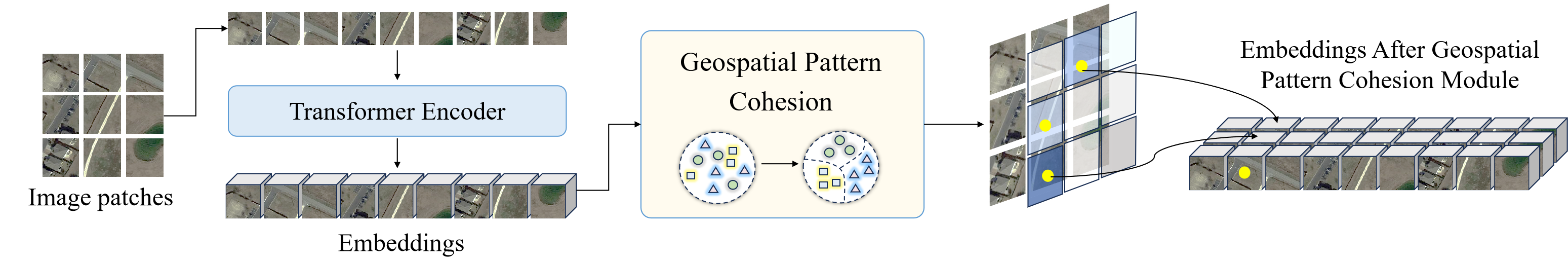}
\caption{Geospatial Pattern Cohesion Score Calculation.}
\label{fig_2}
\end{figure*}

In the realm of remote sensing imagery, captured from an overhead perspective, the landscape unfolds in distinct patterns of spatial distribution. This perspective brings into focus a complex interplay of natural features such as mountains, rivers, forests, and grasslands, alongside human-made elements like buildings, agricultural fields, and urban layouts. These features tend to form discernible clusters, creating a unique spatial relationship between different patches and their surrounding areas. To effectively leverage these spatial patterns and enhance the student network's masked token optimization with a nuanced understanding of geographical contexts, we introduce the GPC Module, as shown in Figure \ref{fig_2}. This module is designed to delve into and exploit the inherent clustering of landscape elements within remote sensing images, thereby significantly improving the model's ability to interpret complex spatial distributions.

Specifically, the GPC Module is applied to query patch embeddings derived from image patches. This module quantifies the GPC score by evaluating the similarity between the embedding of a central patch, denoted as $X_i$, and the embeddings of its adjacent neighboring patches, represented as $X_i^j$. The cohesion score between the central patch and its neighbors is computed to produce a series of cohesion scores:
\begin{equation}
s(i,j)=\frac{X_i^T\cdot X_i^j}{\|X_i\|_2\cdot\|X_i^j\|_2 }\;,
\end{equation}

we select the top $k$ patches ($\{s_{(1)}, s_{(2)}, ..., s_{(k)}$, where $s_{(1)}\geq s_{(2)}\geq ...\geq s_{(k)}$) with the highest scores as the output representing geospatial pattern aggregation in the neighborhood, which then guides the generation of masked patches in the subsequent student network.

This method of neighborhood feature aggregation allows us to effectively highlight features that are prominent or representative within local areas, achieving a pattern of GPC. This, in turn, enables a more precise analysis and understanding of the complexity inherent in remote sensing imagery. Such an approach is crucial for guiding and refining the process of generating masked tokens within the student network.

\subsection{The Feature Integration Projection (FIP)}
The Feature Integration Projection (FIP) module is designed to harness the Geospatial Pattern Cohesion scores, derived within the teacher network, to guide the generation of masked patches in the student network. Given that the conventional projection head, denoted as $h$, comprises 3-layer MLPs with an $L_2$-normalized bottleneck, it falls short in addressing the complexities of multi-patch integration. To overcome this limitation, we introduce the FIP Module, specifically tailored to process patch embeddings subsequent to our ViT transformation.

The FIP structure consists of two primary blocks and an MLP head. Each block is composed of a cross-attention mechanism and a Feed Forward Network (FFN). The cross-attention mechanism is pivotal for distilling pertinent information from an array of patch features. It shares a structural resemblance with self-attention; however, its distinctiveness lies in leveraging attention between a targeted set of embeddings, either classification(cls) or patch, and a series of static patch embeddings, thereby facilitating a focused integration of feature information. For conciseness, we use $H_\theta$ to denote the whole process of the FIP module parameterized by $\theta$ as:

\begin{equation}
H_\theta(\hat{x})=MLP(Blocks(\hat{x}))\;,
\end{equation}

\begin{equation}
Blocks(\hat{x})=FFN(CA(FFN(CA(\hat{x}))))\;,
\end{equation}

where CA denotes the cross-attention operation, $\hat{x}$ denotes the embeddings and the architecture of the cross-attention mechanism is defined by multiple projection matrices. This formulation sets the stage for the computation within the CA residual blocks, enabling a structured and efficient integration of features across the patch embeddings. In terms of computational efficiency, the FIP module consists of a set of 2 blocks of layers (2 CA and 2 FFN). The parameter count of each block is the same as that of SA + FFN layers, both being 1.1M. However, CA is more resource-efficient in terms of memory and computation compared to SA because CA computes attention between the class embedding and the patch embeddings, while SA computes attention between the patch embeddings. Since the patch embeddings have a higher dimension, SA requires higher computational overhead.

\subsection{The Dual-Stream Feature Learning Network}
As outlined in Section \ref{SEC:III-A}, the overarching network is a parallel architecture consisting of two neural networks, commonly referred to as the student network and the teacher network, forming a dual-stream framework. The student network is responsible for learning and updating its weights through standard gradient descent (backpropagation). In contrast, the teacher network's parameters are not updated directly through backpropagation. Instead, they are updated using EMA of the student network's parameters. This means that during subsequent error propagation, the teacher network does not receive gradient updates from the loss function. This stop-gradient mechanism ensures that the teacher network's parameters are updated smoothly and remain relatively stable, which helps prevent oscillations during the training process. The specific formula for EMA is as follows: $\theta_t \leftarrow \lambda \theta_t + (1 - \lambda) \theta_s$, where $\theta_t$ and $\theta_s$ represent the weights of the teacher network and the student network, respectively. $\lambda$ is the EMA decay factor, which we have set to 0.99.
\begin{figure}[htpb]
\centering
\includegraphics[width=\columnwidth]{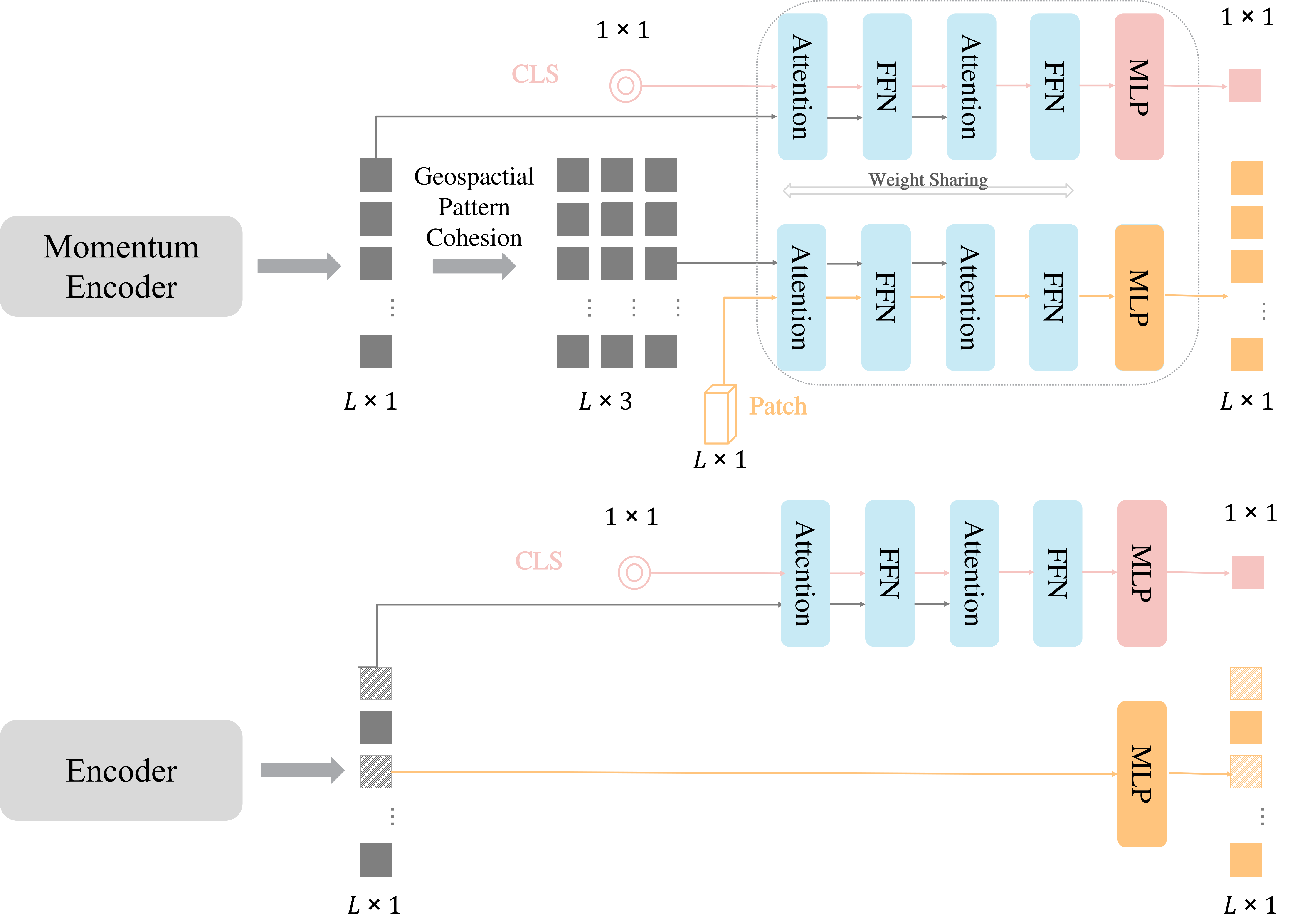}
\caption{The Dual-Stream Feature Learning Framework.}
\label{fig_3}
\end{figure}

Within this structure, the teacher network's architecture adopts tailored strategies at the image and patch levels, as illustrated in the accompanying Figure \ref{fig_3}. For image-level processing, a CLS embedding $X_{class}\in\mathbb{R}^{1\times d}$ is initialized to function as the query features. This embedding is then concatenated with the static patch embeddings $X_{patch}\in\mathbb{R}^{L\times d}$, derived from the ViT, to function as the key and value features. This configuration aligns the CA operation more closely with a Self-Attention (SA) layer by ensuring a corresponding key for every query. At the patch level, an aggregated patch embedding $X_{ap}\in\mathbb{R}^{L\times d}$ is designated as the query features, which is then concatenated with similar, static patch embeddings $X_{sp}\in\mathbb{R}^{L\times 3\times d}$ to serve as key and value features. Consequently, the residual output vectors for CLS and patch embeddings can be articulated as follows:

\begin{equation}
\begin{split}
\textcolor{black}{CA_{cls_{out}}=X_{cls}+Softmax((W_qX_{cls}+b_q)(W_k[X_{cls},}\\ \textcolor{black}{X_{patch}]+b_k)^T/\sqrt{d/h})(W_v[X_{cls}, X_{patch}]+b_v)\;, }  
\end{split}
\end{equation}

\begin{equation}
\begin{split}
\textcolor{black}{CA_{patch_{out}}=X_{ap}+Softmax((W_qX_{ap}+b_q)(W_k[X_{ap},}\\ \textcolor{black}{X_{sp}]+b_k)^T/\sqrt{d/h})(W_v[X_{ap},X_{sp}]+b_v)\;,}
\end{split}
\end{equation}

$W_q, W_k, W_v\in\mathbb{R}^{d\times d}$ and their respective biases $b_q, b_k, b_v\in\mathbb{R}^{d}$ represent the weight matrices and bias vectors used in the calculation of the cross-attention mechanism. $d$ denotes the embedding size. While the block modules at both the image and patch levels share parameters, the parameters in the final MLP layer remain distinct to differentiate the weights associated with the two targets, thereby mitigating the risk of underfitting at the patch level and overfitting at the image level. Empirical evidence suggests that this approach yields superior outcomes compared to the utilization of shared heads \cite{oquab2023dinov2}.

In the student network's branch, the approach at the image level mirrors that of the teacher network, where a CLS embedding is derived from the static masked patch embeddings produced by the ViT and then introduced as a query into the FIP module. At the patch level, aligning with the objective of Masked Image Modeling (MIM), the masked patch embeddings are fed directly into the MLP. Consistent with the teacher network's design, the student network maintains distinct parameters in the terminal MLP layer at both the image and patch levels, ensuring specificity in parameter application across different levels of processing.

This strategy of aggregating neighborhood features serves to underscore features that are either prominent or emblematic within localized regions, facilitating a more refined analysis and comprehension of the intricacies embedded within remote sensing imagery. Such a methodology is instrumental in steering and refining the process of masked token generation within the student network, optimizing the network's learning efficiency and efficacy in handling complex spatial data.

\subsection{Optimization Function of the Network}
During training, PIEViT is designed to synergize with the $\mathcal{L}_{CLS}$ and $\mathcal{L}_{MIM}$ objectives, which encompass: 1) a global objective function applied across two global views to encapsulate global-discriminative information; and 2) a patch-level contrastive objective within the same view to encapsulate local masked information. Given an image $x$, upon which two random augmentations are applied, we derive two distorted views $u$ and $v$. Subsequent random block masking yields their masked counterparts, $\hat{u}$ and $\hat{v}$. These distorted views are processed through a teacher-student framework to obtain predictive categorical distributions from the CLS token and patch tokens. The training objectives for CLS and MIM within PIEViT are delineated as follows:

\begin{equation}
\mathcal{L}_{CLS}(u, v)=-f_{\theta_t}^{cls}(v)^T\log f_{\theta_s}^{cls}(u)\;,
\end{equation}

\begin{equation}
\mathcal{L}_{CLS}(\hat{u}, \hat{v})=-f_{\theta_t}^{cls}(\hat{v})^T\log f_{\theta_s}^{cls}(\hat{u})\;,
\end{equation}

\begin{equation}
\mathcal{L}_{MIM}(u, \hat{u})=-\sum_{i=1}^{N}m_i f_{\theta_t}^{patch}(u)^T\log f_{\theta_s}^{patch}(\hat{u})\;,
\end{equation}

\begin{equation}
\mathcal{L}_{MIM}(v, \hat{v})=-\sum_{i=1}^{N}m_i f_{\theta_t}^{patch}(v)^T\log f_{\theta_s}^{patch}(\hat{v})\;,
\end{equation}

\begin{equation}
\mathcal{L}_{total}=\mathcal{L}_{CLS}(u, v)+\mathcal{L}_{CLS}(\hat{u}, \hat{v})+\mathcal{L}_{MIM}(u, \hat{u})+\mathcal{L}_{MIM}(v, \hat{v})\;,
\end{equation}

where $\mathcal{L}_{CLS}$ denotes the loss function at the global level, capturing class-level predictive accuracy, while $\mathcal{L}_{MIM}$ focuses on the patch level, emphasizing the learning of local features through masked image modeling. $\mathcal{L}_{total}$ represents the aggregate loss, combining both global and local learning objectives to foster a comprehensive understanding of the image content at multiple scales.

\section{Experiments}
In this section, we conducted extensive experiments to evaluate the effectiveness of the proposed self-supervised learning model PIEViT. We compared PIEViT with other self-supervised learning methods based on models with similar sizes across various remote sensing downstream tasks, including object detection, land cover classification, and change detection. All methods were implemented in PyTorch \cite{paszke2017automatic}.

\subsection{Pretraining}
The pretraining of PIEViT was based on the Million-AID dataset \cite{long2021creating}. Million-AID is a new large-scale benchmark dataset comprising nearly one million remote sensing (RS) scene classification images. It encompasses a wide range of semantic classes, totaling 51 scene categories. PIEViT was trained from scratch for 100 epochs with a batch size of 256, distributed across two A800 GPUs using ViT-B/16 as the backbone.

We used the AdamW optimizer with hyperparameters $\beta_1=0.9$ and $\beta_2=0.99$. The learning rate was warmed up linearly to a base value of 1e-4 during the first 10 epochs, followed by a decay according to a cosine schedule. Weight decay also followed a cosine schedule from 0.04 to 0.4. The initial LayerScale value was set to 1e-5, and the teacher momentum followed a cosine schedule from 0.994 to 1. We utilized several data augmentation techniques, including color jittering, Gaussian blur, and multi-crop. Additionally, we performed random MIM when training PIEViT, with the mask ratio randomly chosen between 0.1 and 0.5. After pretraining the PIEViT, we use its weight parameters as the backbone and combine it with downstream task-specific heads for fine-tuning.

We compile the number of parameters and FLOPs (floating point operations) for each model during pre-training. FLOPs are computed for a given size of 224×224. Table \ref{tab:1} provides a detailed computational complexity comparison of the pre-training models.

\begin{table}
    \centering
    \caption{Computational complexity comparison.}
    \begin{tabular}{c|c|c|c}
         \cline{1-4}
         Method&Encoder&Params.&FLOPs \\ \cline{1-4}
         DINO\cite{caron2021emerging}&ViT-B&86M&23.5G \\
         IBOT\cite{zhou2021ibot}&ViT-B&86.58M&23.8G \\
         MAE\cite{he2022masked}&ViT-B&111.91M&17.58G  \\
         SatMAE\cite{cong2022satmae}&ViT-L&329.5M&61.6G \\
         RingMo\cite{sun2022ringmo}&Swin-B&87.75M&11.3G \\
         RVSA\cite{wang2022advancing}&ViT-B&111.91M&17.58G  \\
         GFM\cite{mendieta2023building}&Swin-B&87.75M&11.3G  \\
         Scale-MAE\cite{huang2024generic}&ViT-L&322.9M&61.6G \\
         PIEViT&ViT-B&88.2M&24.2G \\
         \cline{1-4}
    \end{tabular}
    \label{tab:1}
\end{table}

\subsection{Object Detection}
We used the DIOR dataset \cite{li2020object} to conduct experiments for the object detection task. DIOR is a large-scale, publicly available benchmark for object detection in optical remote sensing images, containing 23,463 images and 192,472 instances across 20 object classes. The training, validation, and testing sets comprise 5,862, 5,863, and 11,738 images, respectively.

We evaluated our model using the popular Faster-RCNN \cite{ren2015faster} framework with the ViT backbone pretrained by PIEViT. The ViT backbone was adapted for use with adapter \cite{chen2022vision}. We trained all models for 12 epochs using an AdamW optimizer with a batch size of 6, a learning rate of 1e-3, and a weight decay of 0.05.
Table \ref{tab:2} shows that after fine-tuning Faster-RCNN for 12 epochs, the proposed PIEViT achieved significant improvements in mAP50 over DINO, MAE, and iBOT, which are also self-supervised trained based on Million-AID, increasing by 4.96\%, 5.18\%, and 4.35\%, respectively. Compared to ViT models supervised on ImageNet, PIEViT led by 2.22\% in mAP50. Additionally, PIEViT demonstrated higher accuracy than other self-supervised algorithms in the remote sensing field, achieving a mAP50 score of  76.92\% on the DIOR test datasets, despite its smaller parameter count.

\begin{table}
    \centering
    \caption{Object detection performance comparison.}
    \begin{tabular}{c|c|c|c|c|c}
         \cline{1-6}
         Method&Backbone&Arch&Param&Epoch&mAP50 \\ \cline{1-6}
         Sup. (IN)\cite{russakovsky2015imagenet}&ViT-B&Faster RCNN&86M&12&74.70 \\
         DINO\cite{caron2021emerging}&ViT-B&Faster RCNN&86M&12&71.96 \\
         IBOT\cite{zhou2021ibot}&ViT-B&Faster RCNN&86M&12&71.74 \\
         MAE\cite{he2022masked}&ViT-B&Faster RCNN&86M&12&72.57  \\
         SatMAE\cite{cong2022satmae}&ViT-L&Faster RCNN&307M&12&70.89 \\
         RingMo\cite{sun2022ringmo}&Swin-B&Faster RCNN&88M&12&75.90 \\
         RVSA\cite{wang2022advancing}&ViT-B&Faster RCNN&86M&12&73.22  \\
         GFM\cite{mendieta2023building}&Swin-B&Faster RCNN&88M&12&72.84  \\
         Scale-MAE\cite{huang2024generic}&ViT-L&Faster RCNN&307M&12&73.81 \\
         PIEViT&ViT-B&Faster RCNN&86M&12&\textbf{76.92} \\
         \cline{1-6}
    \end{tabular}
    \label{tab:2}
\end{table}

\subsection{Land Cover Classification}
We conducted experiments for the land cover classification task using the Potsdam dataset \cite{sherrah2016fully}, which includes 38 satellite images at a 0.05-meter resolution, each with a size of 6000 × 6000 pixels. According to the official splitting, 24 images are used for training, and 14 are used for testing. We cropped the original images to patches with 512 × 512 pixels, resulting in 3,456 training and 2,016 testing samples.

We utilized the ViT backbone pretrained by PIEViT and validated the model using the UperNet network \cite{xiao2018unified}. The ViT backbone was adapted to work with adapter. All models were trained for 20,000 iterations using an AdamW optimizer with a batch size of 8, a learning rate of 1e-6, and a weight decay of 0.01.

As shown in Table \ref{tab:3}, PIEViT achieved significant improvement. With the same number of parameters and computational resources, it outperformed DINO, MAE, and iBOT in the mF1 evaluation metric and even surpassed the ImageNet-supervised ViT by 2.76\%. Furthermore, PIEViT outperforms some self-supervised algorithms using hierarchical ViT structures. For example, GFM uses the Swin Transformer \cite{liu2021swin} as its backbone network, but its mF1 accuracy is 0.87\% lower than that of PIEViT.

\begin{table}
    \centering
    \caption{Land cover classification performance comparison.}
    \begin{tabular}{c|c|c|c|c|c}
         \cline{1-6}
         Method&Backbone&Arch&Param&Iter.&mF1 \\ \cline{1-6}
         Sup. (IN)\cite{russakovsky2015imagenet}&ViT-B&UperNet&86M&20K&89.96 \\
         DINO\cite{caron2021emerging}&ViT-B&UperNet&86M&20K&88.01 \\
         IBOT\cite{zhou2021ibot}&ViT-B&UperNet&86M&20K&88.35 \\
         MAE\cite{he2022masked}&ViT-B&UperNet&86M&20K&89.44  \\
         SatMAE\cite{cong2022satmae}&ViT-L&UperNet&307M&160K&90.63 \\
         RingMo\cite{sun2022ringmo}&Swin-B&UperNet&88M&160K&91.27 \\
         GFM\cite{mendieta2023building}&Swin-B&UperNet&88M&160K&91.85  \\
         Scale-MAE\cite{huang2024generic}&ViT-L&UperNet&307M&160K&91.54 \\
         PIEViT&ViT-B&UperNet&86M&20K&\textbf{92.72} \\
         \cline{1-6}
    \end{tabular}
    \label{tab:3}
\end{table}

\subsection{Change Detection}
We used the LevirCD dataset \cite{chen2020spatial} for change detection experiments. LEVIR-CD is a widely used building change detection dataset containing 637 very high-resolution (0.5 m) Google Earth image pairs, each with a size of 1024 × 1024 pixels. The training, validation, and testing sets consist of 445, 64, and 128 pairs, respectively. We cropped the images to 512 × 512 pixels, resulting in 1,189 training, 168 validation, and 353 testing samples.

We used the ViT backbone pretrained by PIEViT and validated the model using the BIT \cite{chen2021remote}. The ViT backbone was adapted for use with adapter. All models were trained for 100 epochs with an AdamW optimizer using a batch size of 8, a learning rate of 1e-4, and a weight decay of 0.05.

Table \ref{tab:4} shows the change detection performance on the LevirCD test datasets. Under similar parameter counts and computational resources, PIEViT outperformed DINO, iBOT, and MAE in the mF1 evaluation metric. Compared to the ViT supervised on ImageNet, PIEViT achieved an additional 1.72\% improvement in mF1. PIEViT’s performance was only slightly lower than that of Scale-MAE, which has more than three times the parameters of PIEViT.

\begin{table}
    \centering
    \caption{Change detection performance comparison.}
    \begin{tabular}{c|c|c|c|c|c}
         \cline{1-6}
         Method&Backbone&Arch&Param&Epoch&mF1 \\ \cline{1-6}
         Sup. (IN)\cite{russakovsky2015imagenet}&ViT-B&BIT&86M&100&90.22 \\
         DINO\cite{caron2021emerging}&ViT-B&BIT&86M&100&89.14 \\
         IBOT\cite{zhou2021ibot}&ViT-B&BIT&86M&100&89.17 \\
         MAE\cite{he2022masked}&ViT-B&BIT&86M&100&89.62  \\
         SatMAE\cite{cong2022satmae}&ViT-L&BIT&307M&200&87.65 \\
         RingMo\cite{sun2022ringmo}&Swin-B&BIT&88M&200&91.86 \\
         GFM\cite{mendieta2023building}&Swin-B&BIT&88M&200&91.73  \\
         Scale-MAE\cite{huang2024generic}&ViT-L&BIT&307M&200&\textbf{92.07} \\
         PIEViT&ViT-B&BIT&86M&100&91.94 \\
         \cline{1-6}
    \end{tabular}
    \label{tab:4}
\end{table}

\subsection{Qualitative Analysis}
To assess the quality of PIEViT's feature representation, we conducted several qualitative analyses, including examinations of object detection, semantic segmentation, and change detection in downstream tasks, as well as an analysis of the patch features extracted by PIEViT.

\subsubsection{Object Detection Results}
\begin{figure*}[htbp]
    \centering
    \begin{subfigure}[b]{0.11\textwidth}
        \includegraphics[width=\textwidth]{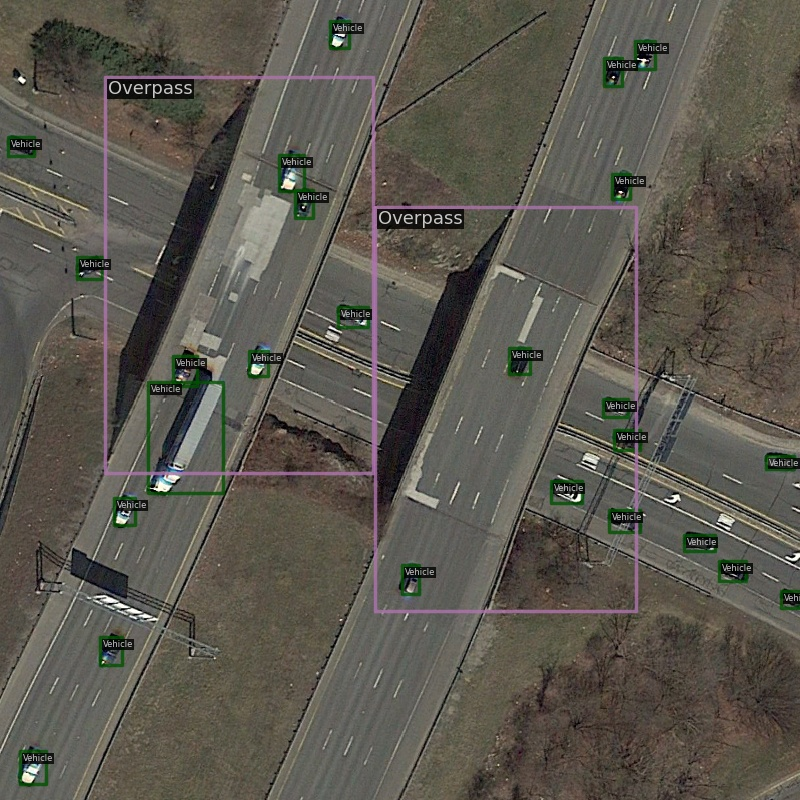}
    \end{subfigure}
    \hfill
    \begin{subfigure}[b]{0.11\textwidth}
        \includegraphics[width=\textwidth]{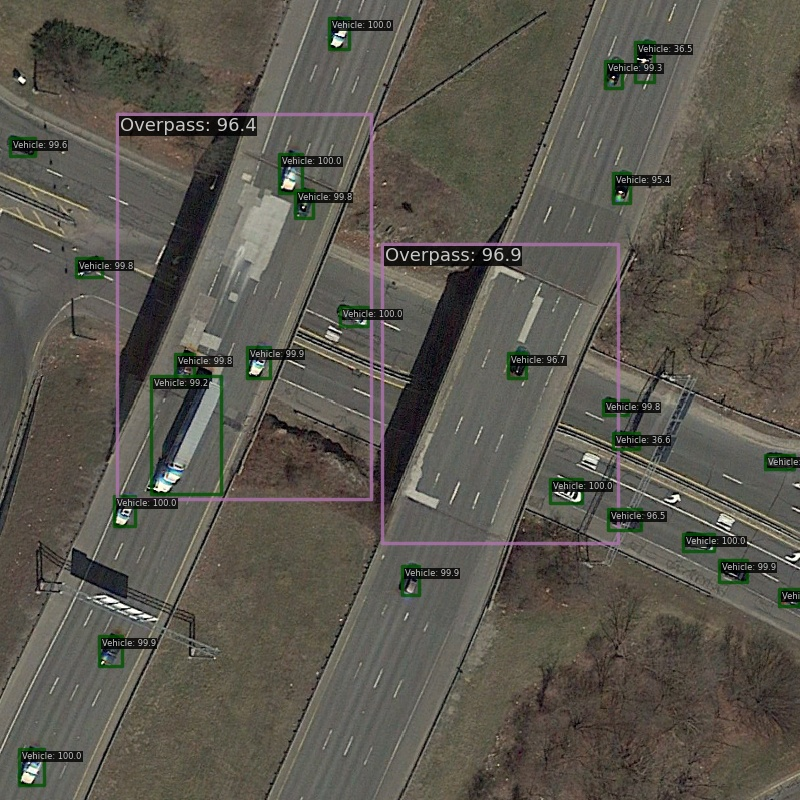}
    \end{subfigure}
    \hfill
    \begin{subfigure}[b]{0.11\textwidth}
        \includegraphics[width=\textwidth]{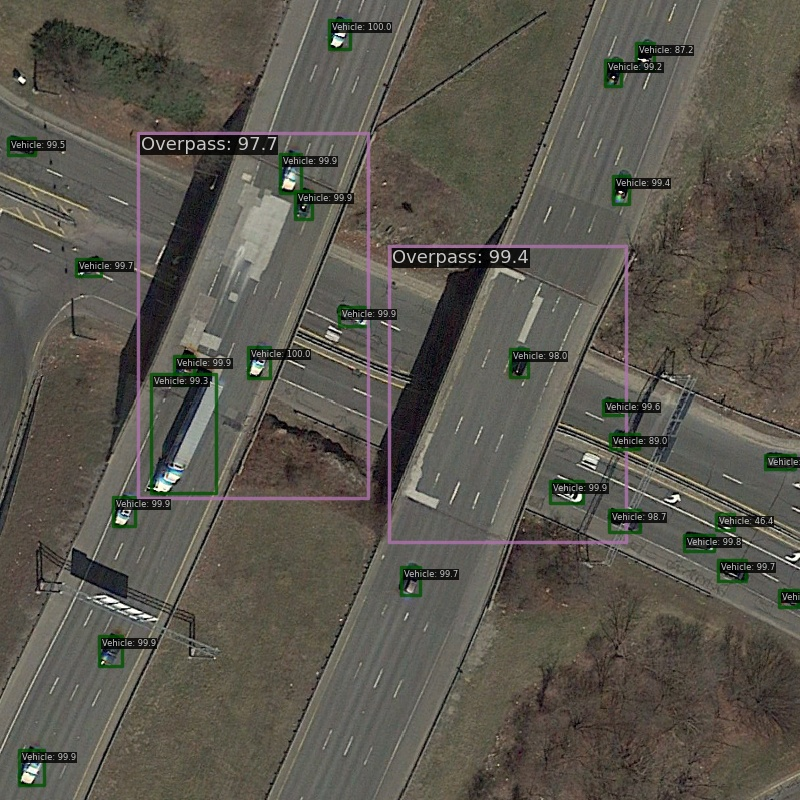}
    \end{subfigure}
    \hfill
    \begin{subfigure}[b]{0.11\textwidth}
        \includegraphics[width=\textwidth]{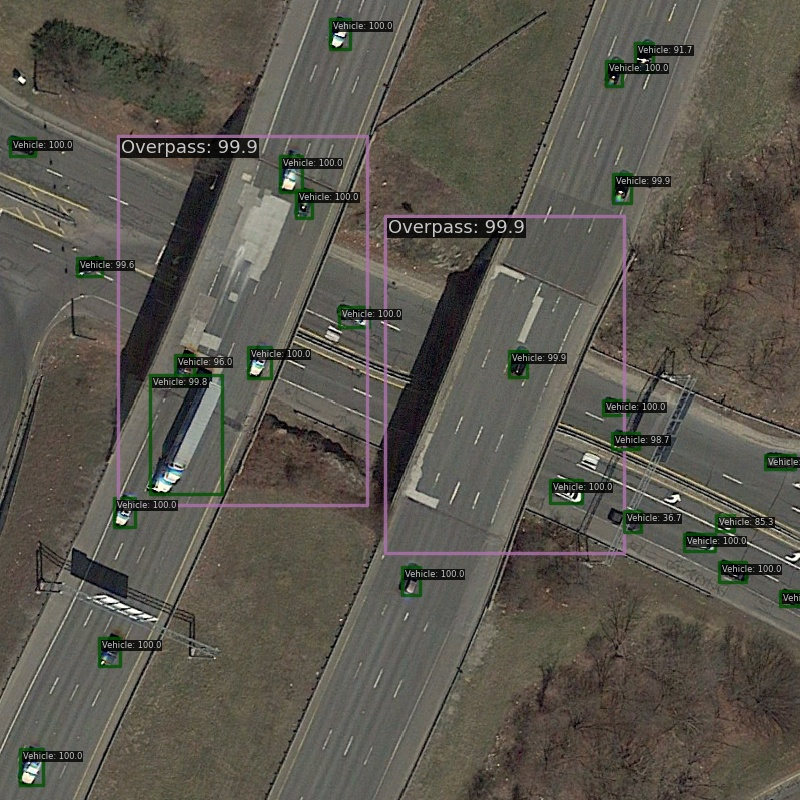}
    \end{subfigure}
    \hfill
    \begin{subfigure}[b]{0.11\textwidth}
        \includegraphics[width=\textwidth]{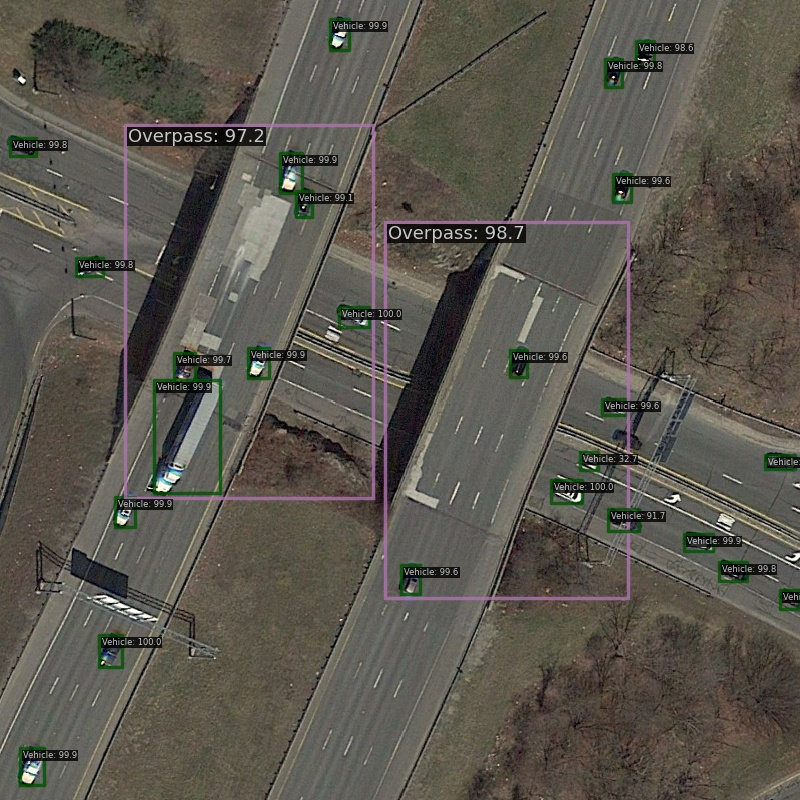}
    \end{subfigure}
    \hfill
    \begin{subfigure}[b]{0.11\textwidth}
        \includegraphics[width=\textwidth]{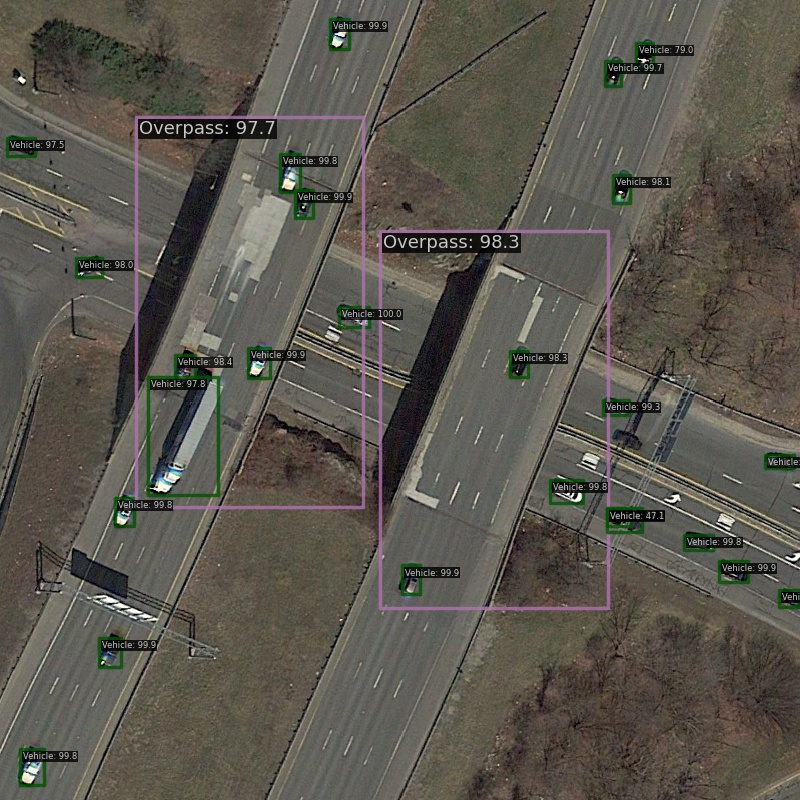}
    \end{subfigure}
    \hfill
    \begin{subfigure}[b]{0.11\textwidth}
        \includegraphics[width=\textwidth]{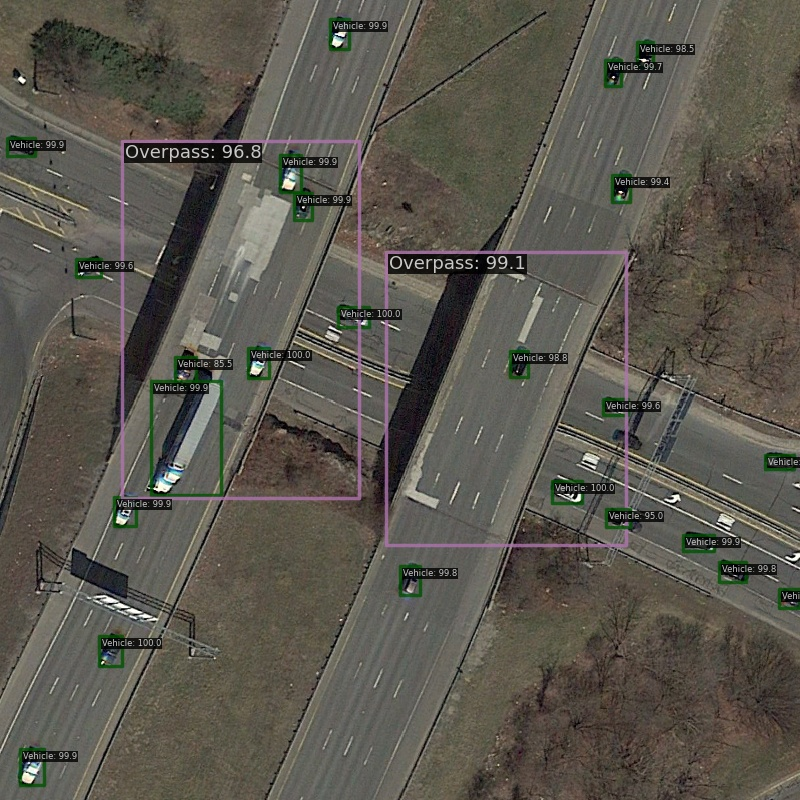}
    \end{subfigure}
    \hfill
    \begin{subfigure}[b]{0.11\textwidth}
        \includegraphics[width=\textwidth]{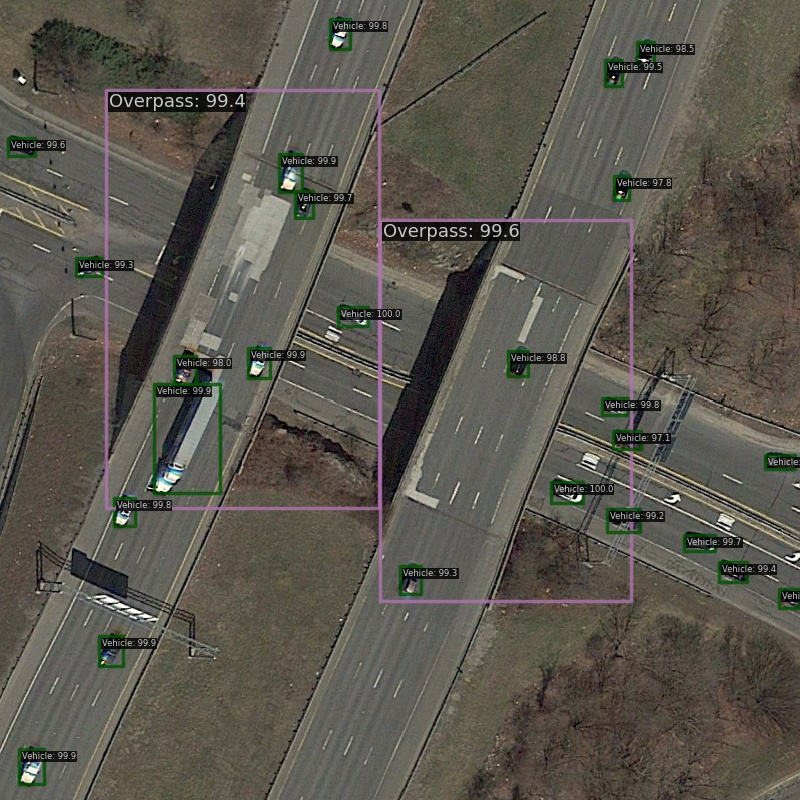}
    \end{subfigure}

    \begin{subfigure}[b]{0.11\textwidth}
        \includegraphics[width=\textwidth]{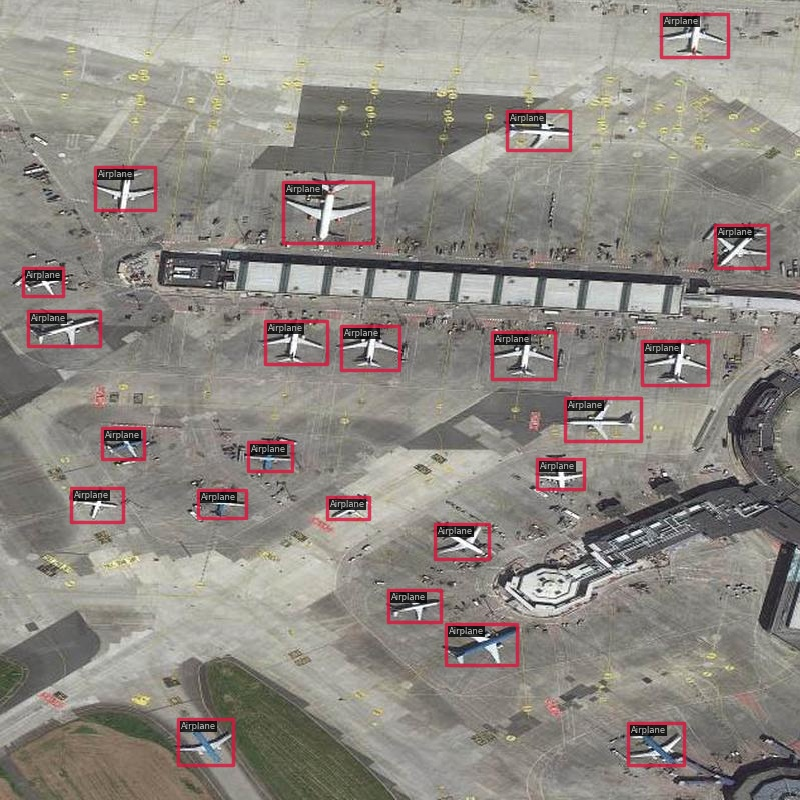}
    \end{subfigure}
    \hfill
    \begin{subfigure}[b]{0.11\textwidth}
        \includegraphics[width=\textwidth]{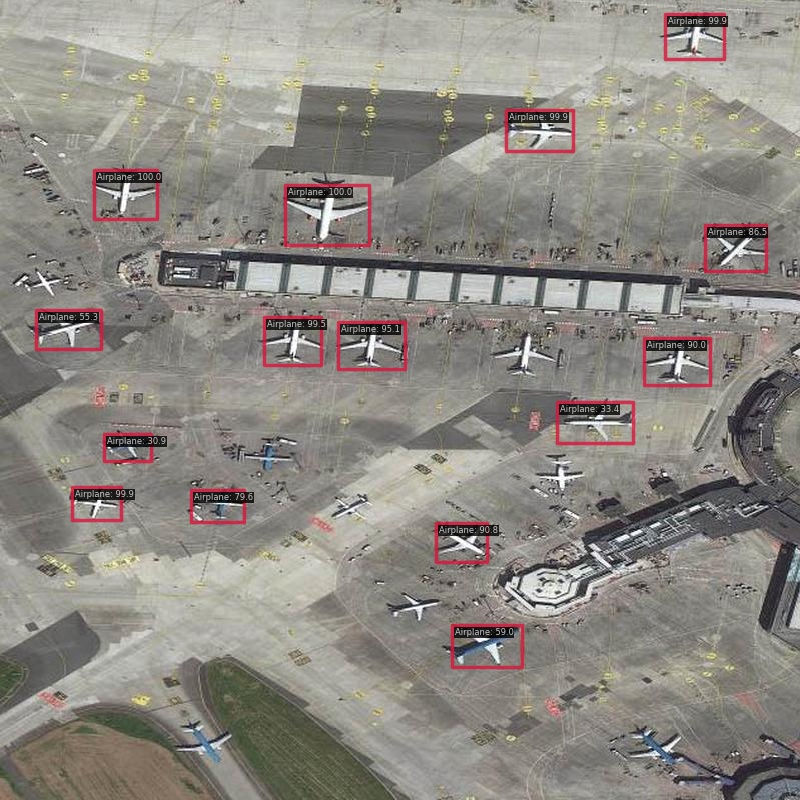}
    \end{subfigure}
    \hfill
    \begin{subfigure}[b]{0.11\textwidth}
        \includegraphics[width=\textwidth]{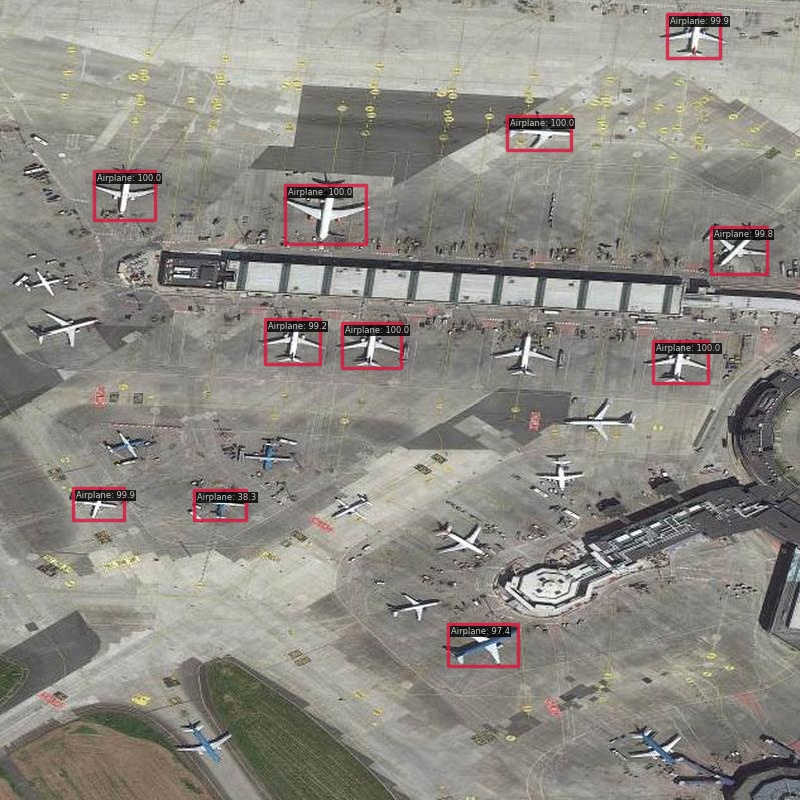}
    \end{subfigure}
    \hfill
    \begin{subfigure}[b]{0.11\textwidth}
        \includegraphics[width=\textwidth]{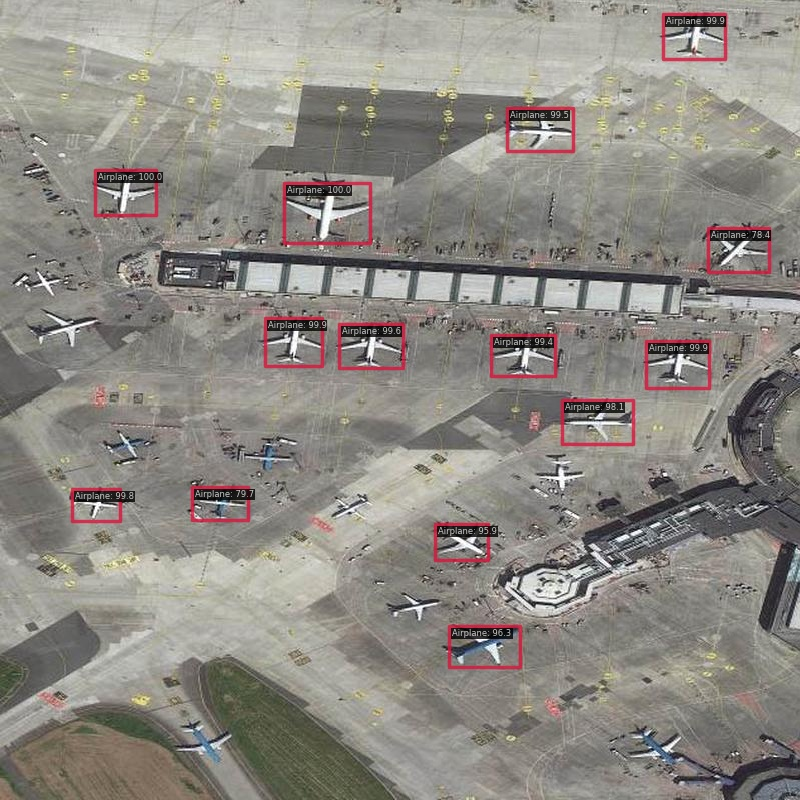}
    \end{subfigure}
    \hfill
    \begin{subfigure}[b]{0.11\textwidth}
        \includegraphics[width=\textwidth]{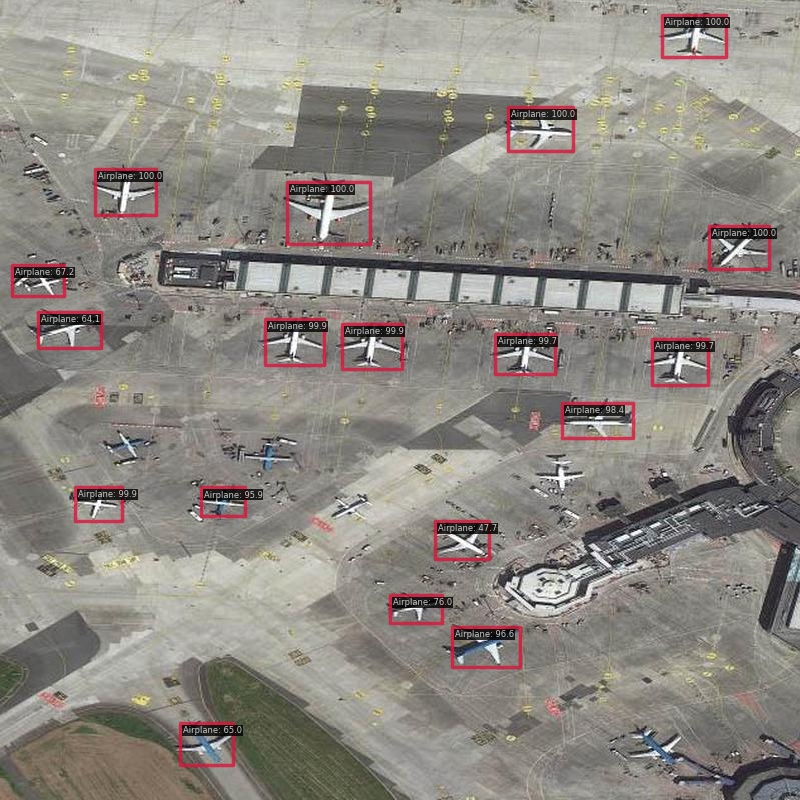}
    \end{subfigure}
    \hfill
    \begin{subfigure}[b]{0.11\textwidth}
        \includegraphics[width=\textwidth]{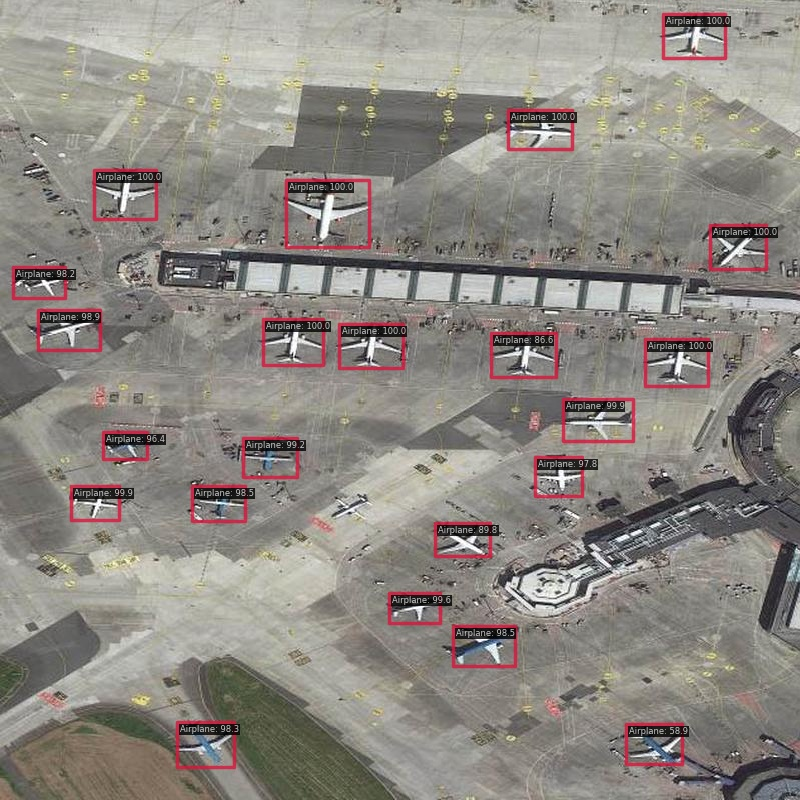}
    \end{subfigure}
    \hfill
    \begin{subfigure}[b]{0.11\textwidth}
        \includegraphics[width=\textwidth]{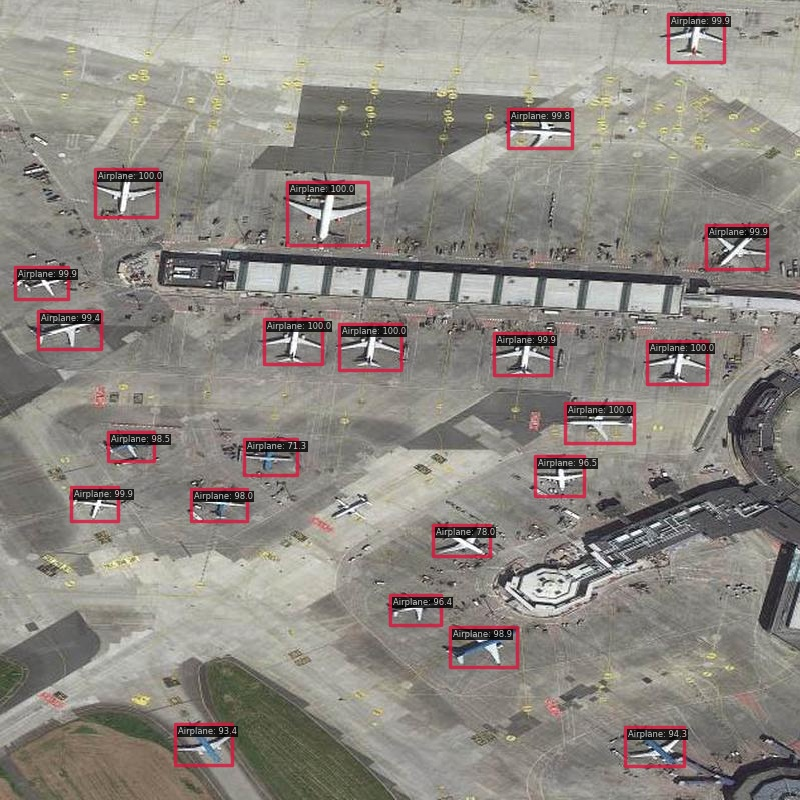}
    \end{subfigure}
    \hfill
    \begin{subfigure}[b]{0.11\textwidth}
        \includegraphics[width=\textwidth]{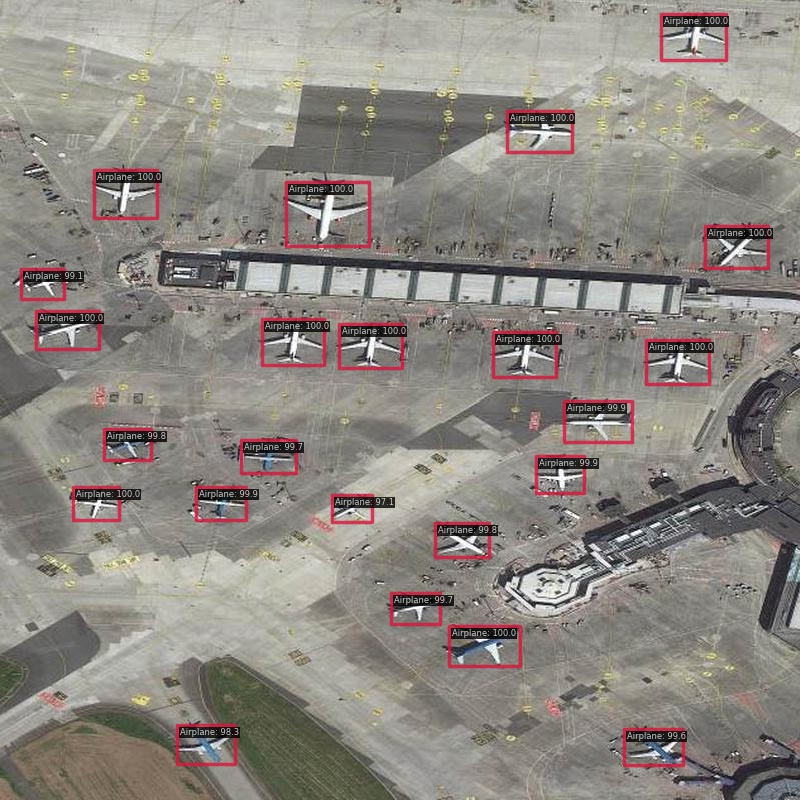}
    \end{subfigure}

    \begin{subfigure}[b]{0.11\textwidth}
        \includegraphics[width=\textwidth]{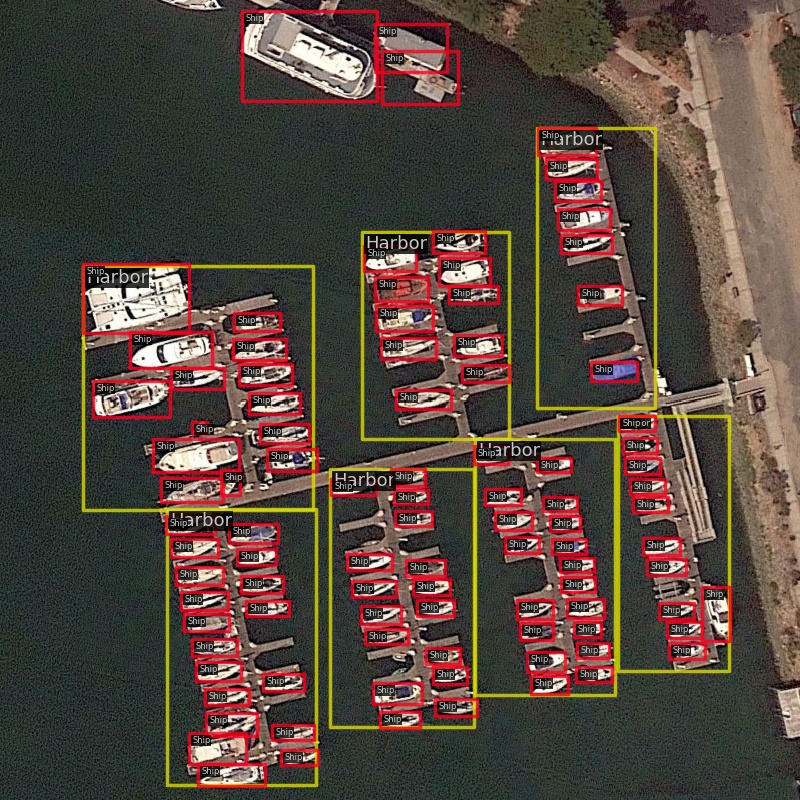}
        \caption{GT}
    \end{subfigure}
    \hfill
    \begin{subfigure}[b]{0.11\textwidth}
        \includegraphics[width=\textwidth]{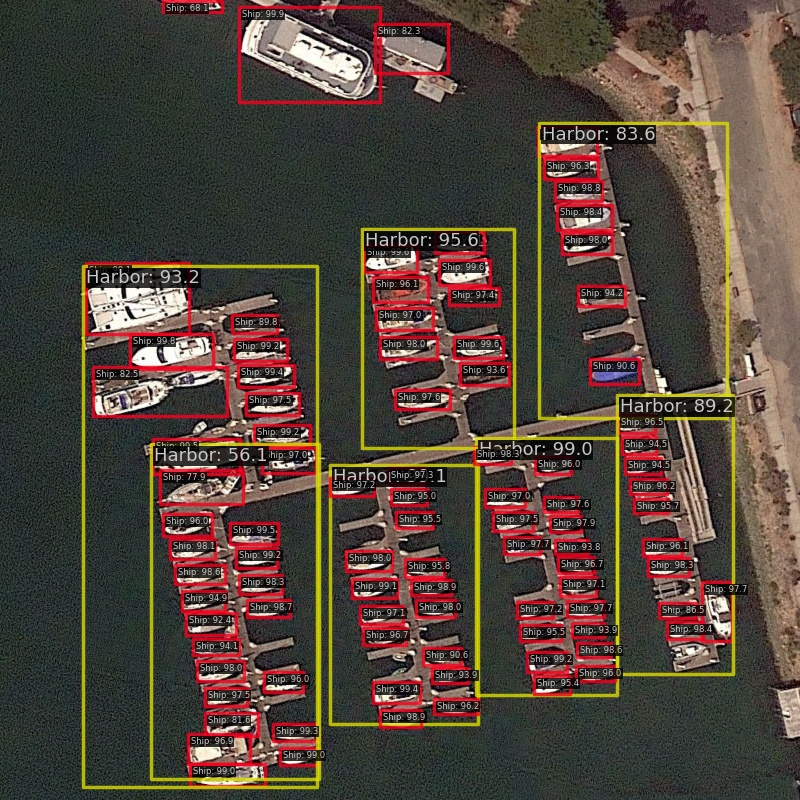}
        \caption{Sup. (IN)}
    \end{subfigure}
    \hfill
    \begin{subfigure}[b]{0.11\textwidth}
        \includegraphics[width=\textwidth]{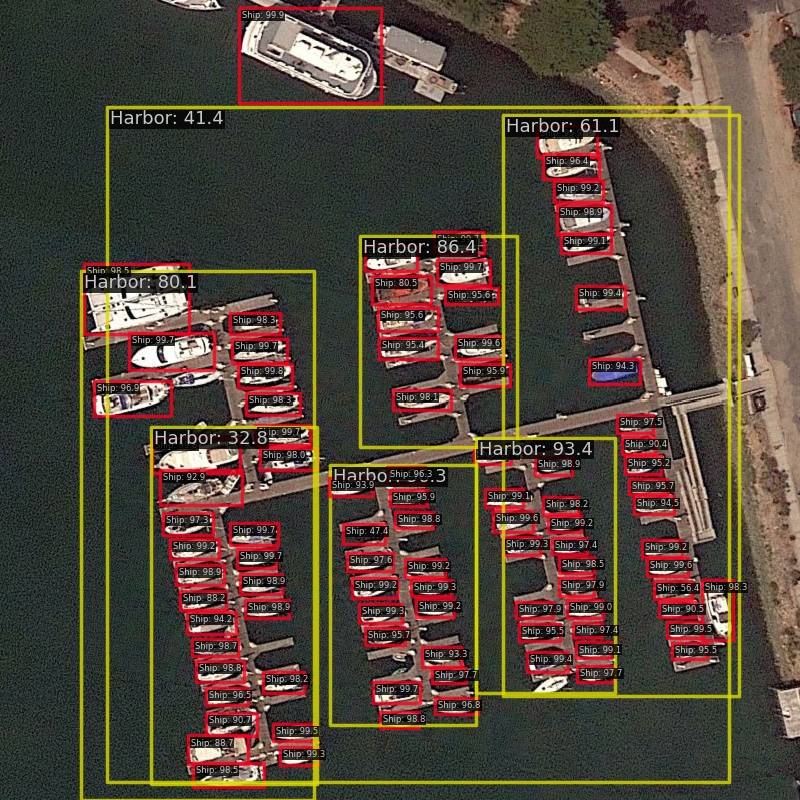}
        \caption{DINO}
    \end{subfigure}
    \hfill
    \begin{subfigure}[b]{0.11\textwidth}
        \includegraphics[width=\textwidth]{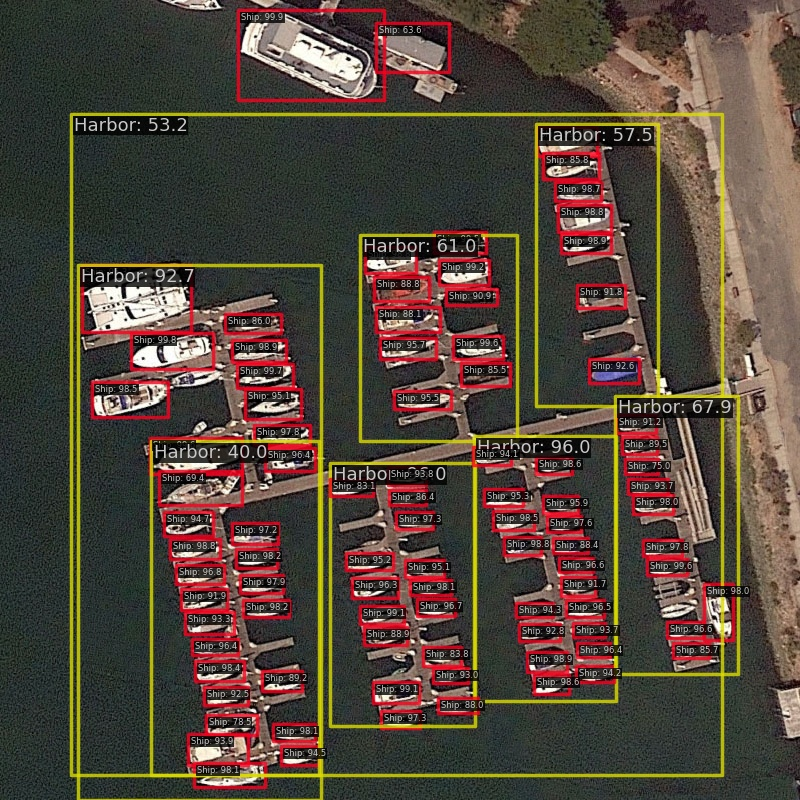}
        \caption{IBOT}
    \end{subfigure}
    \hfill
    \begin{subfigure}[b]{0.11\textwidth}
        \includegraphics[width=\textwidth]{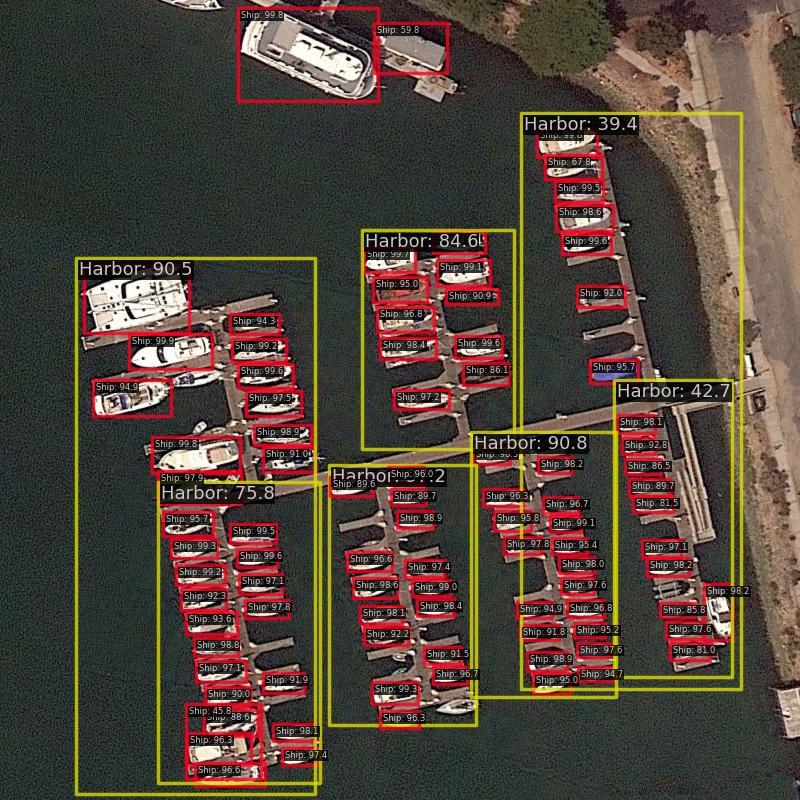}
        \caption{MAE}
    \end{subfigure}
    \hfill
    \begin{subfigure}[b]{0.11\textwidth}
        \includegraphics[width=\textwidth]{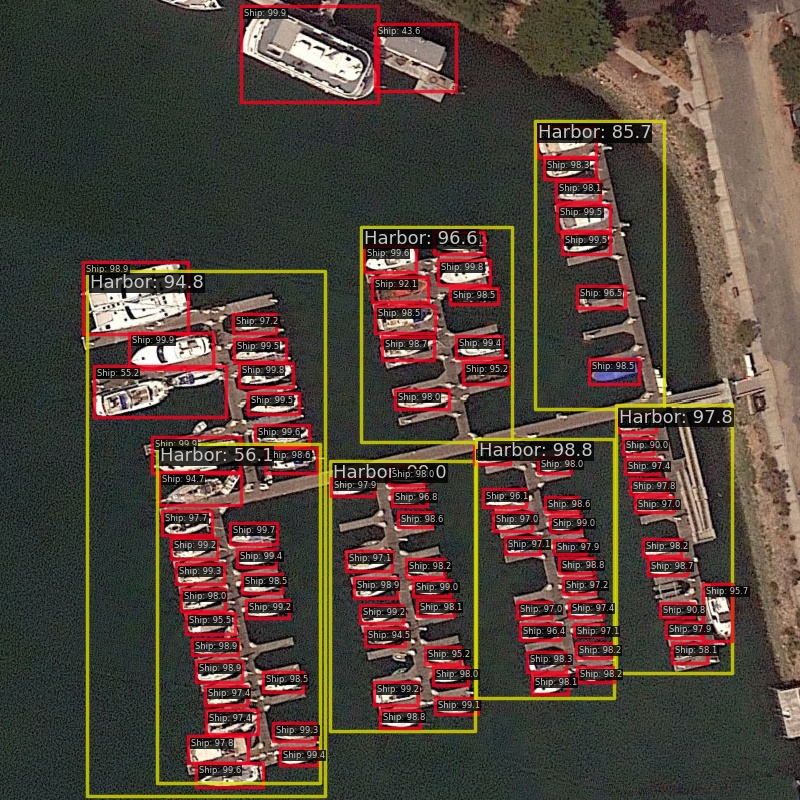}
        \caption{SatMAE}
    \end{subfigure}
    \hfill
    \begin{subfigure}[b]{0.11\textwidth}
        \includegraphics[width=\textwidth]{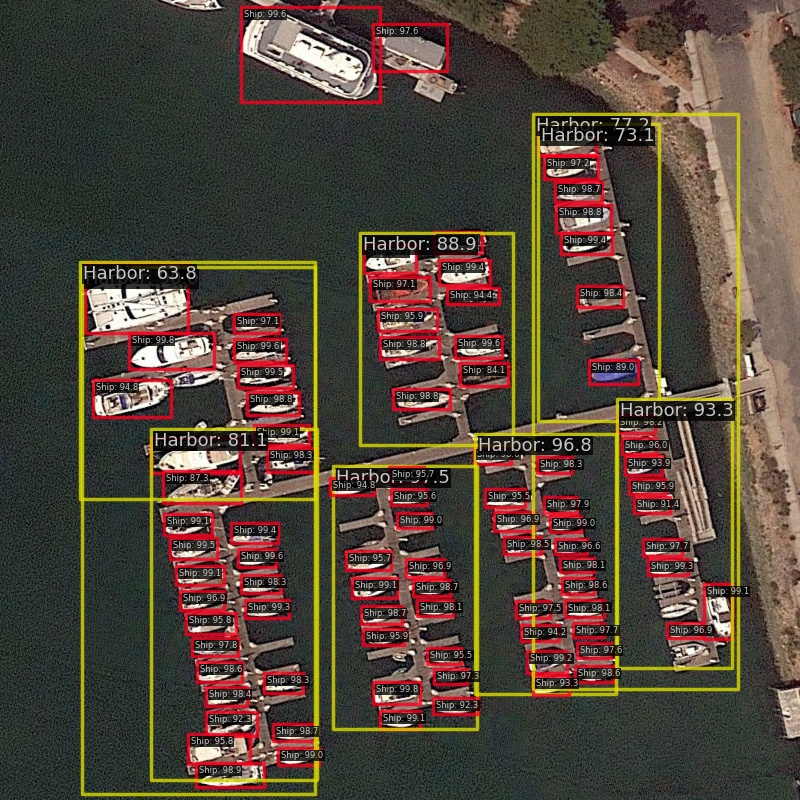}
        \caption{Scale-MAE}
    \end{subfigure}
    \hfill
    \begin{subfigure}[b]{0.11\textwidth}
        \includegraphics[width=\textwidth]{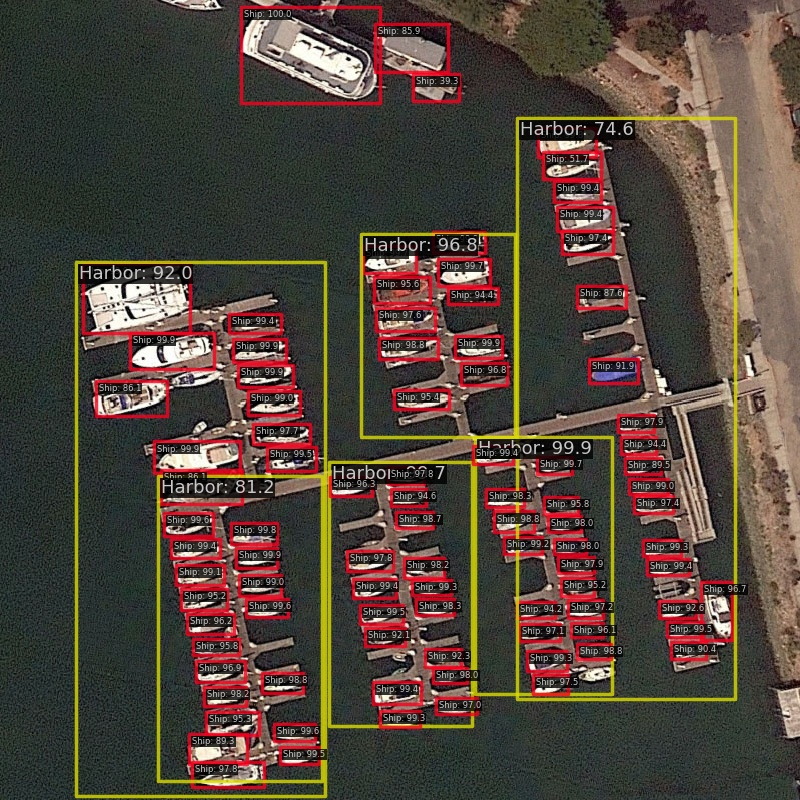}
        \caption{PIEViT}
    \end{subfigure}
    \caption{Visualization of object detection results on the DIOR test dataset.}
    \label{fig_4}
\end{figure*}

Figure \ref{fig_4} provides a visual comparison of different pretraining methods on the DIOR test set. The first row shows an airport scene where models like DINO, iBOT, and MAE missed detecting many airplanes. SatMAE and Scale-MAE could effectively reduce these errors, but still had difficulty when the airplanes closely matched the background in color. Our method detected all airplanes with high confidence. The second row features an overpass scene where the task is to detect both the overpass and vehicles of varying sizes. SatMAE and Scale-MAE had cases of missed and erroneous detections, with SatMAE missing a black car and Scale-MAE misidentifying a white traffic marker as a car. MAE performed well, but its bounding boxes for the overpass and some cars were not accurate. Our method provided bounding boxes that accurately fit the targets' boundaries. The third row shows a complex dock with ships of varying sizes densely arranged. Our method detected every target without redundant bounding boxes, a problem seen with other methods. These results clearly show that PIEViT surpasses other methods, demonstrating the transferability of its feature representation in object detection tasks.

\captionsetup[subfigure]{labelformat=empty}

\subsubsection{Land Cover Classification Results}
\begin{figure*}[htpb]
\centering
\includegraphics[width=\textwidth]{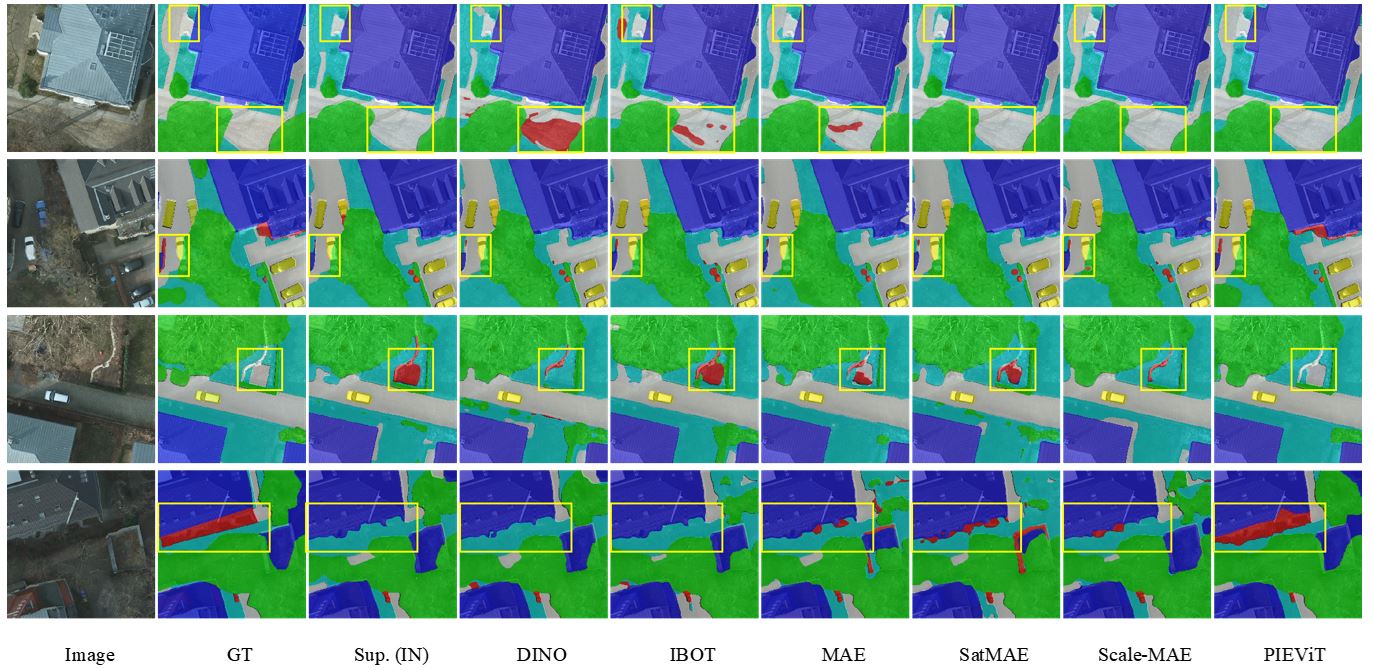}
\caption{Visualization of semantic segmentation results on the Potsdam validation dataset.}
\label{fig_5}
\end{figure*}

Figure \ref{fig_5} illustrates the results of different pretraining methods on the Potsdam validation set. The labels represent impervious surfaces in white, buildings in blue, low vegetation in cyan, trees in green, cars in yellow, and clutter/background in red. Other methods, including SatMAE and Scale-MAE, struggled to distinguish similar features, such as the confusion between impervious surfaces and low vegetation in the top-left corner of the first row. Furthermore, other methods could not effectively separate land cover from the background, misidentifying the background as trees or buildings or mistaking impervious surfaces for the background. In contrast, our method accurately identified these features, avoiding confusion between land covers or the background, demonstrating PIEViT's superior transferability for land cover classification.

\subsubsection{Change Detection Results}
\begin{figure*}[htpb]
\centering
\includegraphics[width=\textwidth]{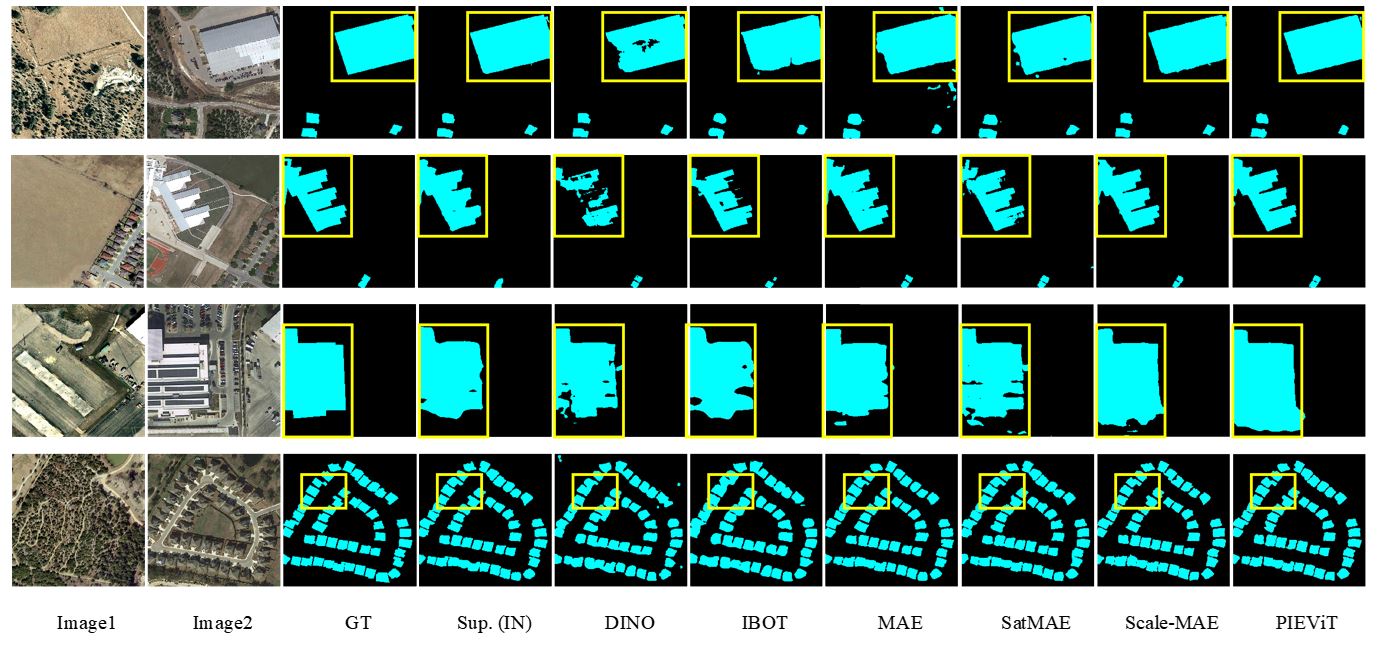}
\caption{Visualization of change detection results on the LevirCD test dataset.}
\label{fig_6}
\end{figure*}

Figure \ref{fig_6} compares different pretraining methods on the LevirCD test dataset, showing changes or additions in buildings. Overall, our method and Scale-MAE produced the best visual results, while other methods struggled with issues such as holes or broken boundaries. Compared to Scale-MAE, PIEViT's boundaries align better with the ground truth, providing the most accurate change map. These results indicate that PIEViT's feature representation is highly transferable and generalized for change detection tasks.

\subsubsection{PCA of Patch Features}
\begin{figure*}[htbp]
    \centering
    \begin{subfigure}[b]{1.0\textwidth}
        \includegraphics[width=\textwidth]{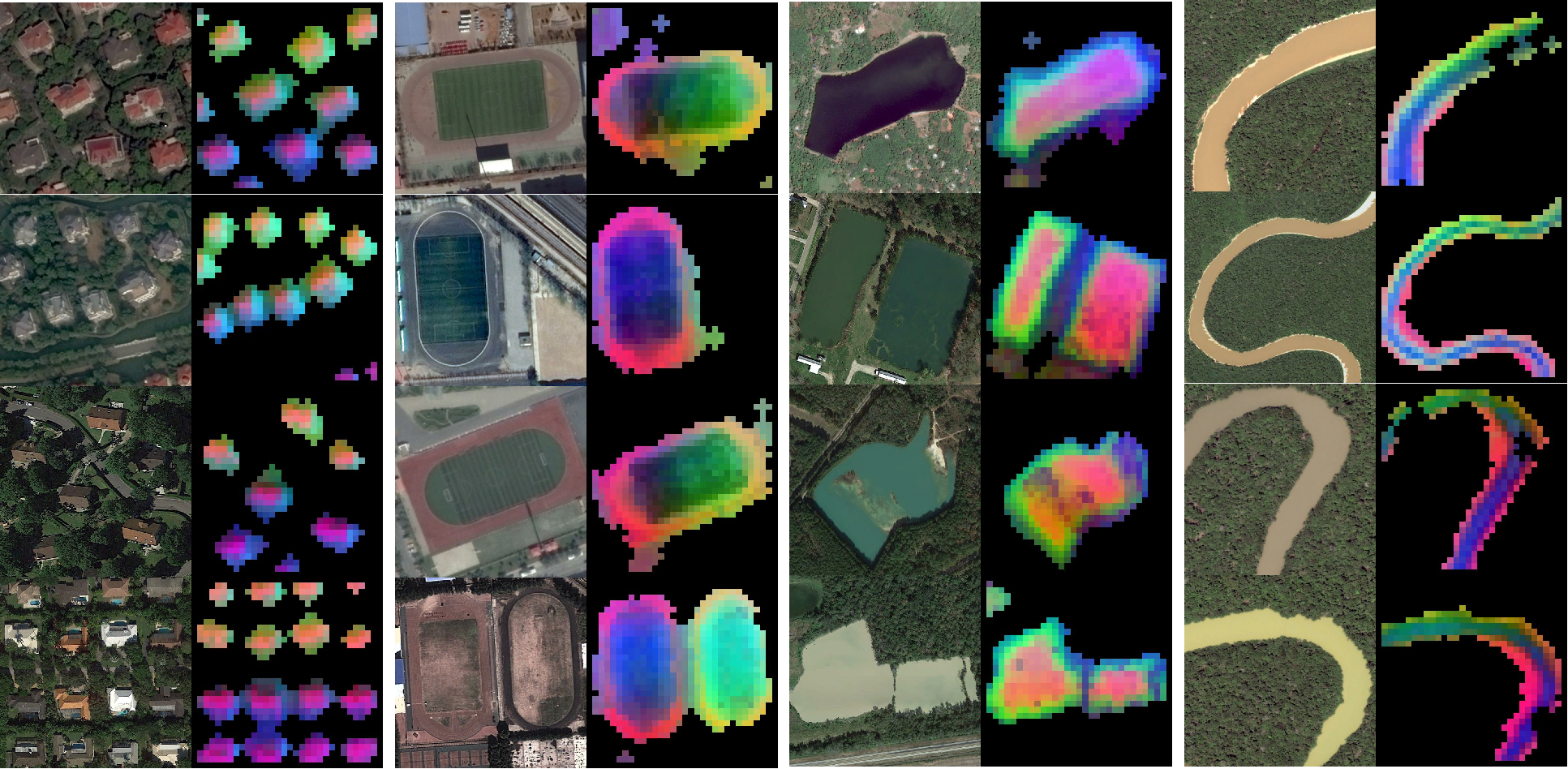}
    \end{subfigure}
    \\[0.3cm] 
    \begin{subfigure}[b]{1.0\textwidth}
        \includegraphics[width=\textwidth]{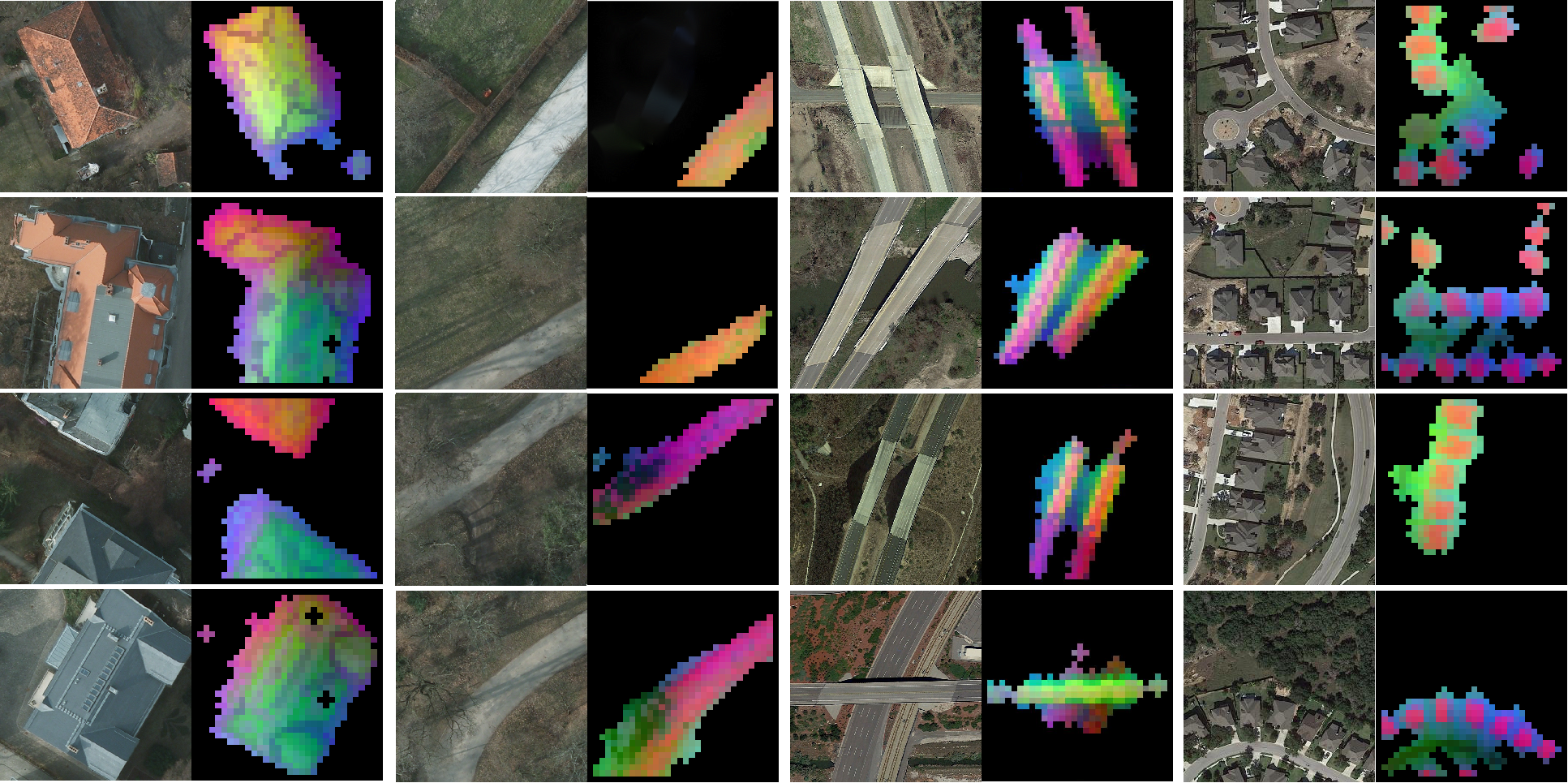}
    \end{subfigure}
    \vfill
    \caption{Visualization of the PCA components. Each column represents images of the same object class, with original images at the left and the PCA results at the right. We visualize the PCA with the first three principal components. Each principal component is rendered with a specific color channel.}
    \label{fig_7}
\end{figure*}

To further verify the feature representation capabilities of PIEViT, we selected four images representing the same category from the Million AID test dataset, DIOR test dataset, Potsdam validation dataset, and LevirCD test dataset, and conducted a Principal Component Analysis (PCA) \cite{abdi2010principal} on the patch features extracted by PIEViT. First, we computed the first PCA  on all patches, applying a 0.5 threshold to filter out the first component, thus separating the foreground from the background and retaining the primary objects. Next, we calculated a second PCA on the remaining patches and visualized the top three components with different colors, as shown in Figure \ref{fig_7}. Notably, PIEViT demonstrated a strong ability to capture features of the same type of objects. For example, in the far-left image, PIEViT accurately extracted buildings situated within the forest, with highly precise boundaries. Furthermore, the local features of objects within the same category were well-matched, as seen in the football field section of the athletics stadium and the water body in the pond, indicating that the model had been trained to interpret parts of objects. This ability is attributed to PIEViT's focus on Geospatial Pattern Cohesion, which is crucial for its outstanding performance in downstream dense tasks, particularly in complex remote sensing scenarios where objects are often obstructed or scattered.


\section{Ablation Study}
\subsection{FIP vs AVG}
We used the FIP module for multi-patch integration, producing comprehensive features that supervise the generation of masked patches. To verify the effectiveness of the FIP module, we replaced it with an average module (AVG), which simply sums and averages multiple patches to supervise masked patch generation. As shown in Table \ref{tab:5}, PIEViT-FIP outperformed PIEViT-AVG in all three downstream tasks—object detection, semantic segmentation, and change detection—by 3.84\% (mAP50), 3.21\% (mF1), and 1.92\% (mF1), respectively. The results demonstrate that the FIP module, based on the attention mechanism, handles multi-patch features more effectively and achieves accurate neighborhood representations.

\begin{table}
    \centering
    \caption{The impact of the FIP and AVG.}
    \setlength\tabcolsep{2.0mm}
    \begin{tabular}{p{1.5cm}|p{1.1cm}|p{0.5cm}p{0.7cm}|p{0.5cm}p{0.5cm}|p{0.5cm}p{0.5cm}}
         \cline{1-8}
         &&Dior&&Potsdam&&LevirCD& \\ \cline{1-8}
         Method&Backbone&mAP&mAP50&mIoU&mF1&mIoU&mF1 \\ \cline{1-8}
         PIEViT-AVG&	ViT-B&	51.59&	73.08&	79.32&	89.51&	81.86& 90.02 \\
         PIEViT&	ViT-B&	53.23&	76.92&	82.33&	92.72&	84.97&	91.94 \\
         \cline{1-8}
    \end{tabular}
    \label{tab:5}
\end{table}

\subsection{Neighborhood Range}
To investigate the impact of neighborhood range on model performance, we evaluated three neighborhood ranges (3$\times$3, 5$\times$5, and 7$\times$7) when selecting the most similar patches based on cosine similarity. As shown in Table \ref{tab:6}, the model's performance declines as the neighborhood range expands. When the neighborhood range is 7$\times$7, accuracy significantly drops, such as a decrease in mAP50 from 76.92 to 66.41 on the object detection task. This result indicates that using nearby patches is more suitable for remote sensing scenarios than employing patches at greater distances.

\begin{table}
    \centering
    \caption{Ablation studies on varying the neighborhood range for PIEViT.}
    \setlength\tabcolsep{2.0mm}
    \begin{tabular}{p{1.5cm}|p{1.1cm}|p{0.5cm}p{0.7cm}|p{0.5cm}p{0.5cm}|p{0.5cm}p{0.5cm}}
         \cline{1-8}
         &&Dior&&Potsdam&&LevirCD& \\ \cline{1-8}
         Method&Backbone&mAP&mAP50&mIoU&mF1&mIoU&mF1 \\ \cline{1-8}
         PIEViT-5x5&	ViT-B&	47.66&	70.09&	75.65&	86.27&	80.92&	89.57\\
PIEViT-7x7&	ViT-B	&44.13&	66.41&	74.09&	85.16&	79.85&	88.82\\
PIEViT&	ViT-B&	53.23&	76.92&	82.33&	92.72&	84.97&	91.94\\
         \cline{1-8}
    \end{tabular}
    \label{tab:6}
\end{table}

\subsection{Top-k}
We selected the top $k$ patches based on cosine similarity among neighboring patches. To verify the effectiveness of this step, we tested various patch counts with $k\in \{2, 3, 4\}$. For a 3×3 neighborhood range, each patch can have up to 8 neighboring patches. For patches located at the corners of the image, we use the 3 surrounding patches. For patches on the edges, we select the most similar 3 patches from the 5 neighboring patches. When $k=4$, for patches at the corners, the fourth patch was replaced by the one with the highest geospatial pattern cohesion score within the neighborhood range. As shown in Table \ref{tab:7}, selecting 3 patches from the neighborhood resulted in the best performance compared to choosing 2 or 4 patches. This result indicates that the number of patches needs to be moderate. Too few patches lead to inaccurate comprehensive features, while too many patches introduce noise into the aggregated features.

\begin{table}
    \centering
    \caption{Effect of the number of patches for PIEViT.}
    \begin{tabular}{p{1.5cm}|p{1.1cm}|p{0.5cm}p{0.7cm}|p{0.5cm}p{0.5cm}|p{0.5cm}p{0.5cm}}
         \cline{1-8}
         &&Dior&&Potsdam&&LevirCD& \\ \cline{1-8}
         Method&Backbone&mAP&mAP50&mIoU&mF1&mIoU&mF1 \\ \cline{1-8}
PIEViT-2	&ViT-B	&50.92	&72.24&	76.01&	86.53&	82.11&	90.20 \\
PIEViT-4&	ViT-B&	51.66&	73.26&	78.13&	88.27&	82.63&	90.54 \\
PIEViT&	ViT-B&	53.23&	76.92&	82.33&	92.72&	84.97&	91.94 \\

         \cline{1-8}
    \end{tabular}
    \label{tab:7}
\end{table}

\subsection{GPC}
The GPC module exploits the inherent clustering of landscape elements within remote sensing images. To investigate its significance, we conducted an experiment where the GPC module was removed, and the embeddings from the vision encoder were fed directly into the FIP module. As shown in Table \ref{tab:8}, the removal of the GPC module resulted in decreases in performance across three downstream tasks: mAP50 by 5.05\%, mF1 by 6.63\%, and mF1 by 2.06\%. These results indicate that the GPC module is crucial for the model's transfer performance. Additionally, the GPC module's computations involve only dot products and L2 norms, which are computationally efficient and do not significantly affect overall training efficiency.

\begin{table}
    \centering
    \caption{The impact of Geospatial Pattern Cohesion (GPC) on model transfer performance.}
    \setlength\tabcolsep{2.0mm}
    \begin{tabular}{p{1.5cm}|p{1.1cm}|p{0.5cm}p{0.7cm}|p{0.5cm}p{0.5cm}|p{0.5cm}p{0.5cm}}
         \cline{1-8}
         &&Dior&&Potsdam&&LevirCD& \\ \cline{1-8}
         Method&Backbone&mAP&mAP50&mIoU&mF1&mIoU&mF1 \\ \cline{1-8}
         PIEViT&	ViT-B&	53.23&	76.92&	82.33&	92.72&	84.97&	91.94 \\
         --GPC&	ViT-B&	50.17&	71.87&	74.93&	86.09&	81.64&	89.88 \\
         \cline{1-8}
    \end{tabular}
    \label{tab:8}
\end{table}

\section{Conclusion}
In this work, we presented PIEViT, a Pattern Integration and Enhancement Vision Transformer specifically designed for remote sensing applications. Our method uses geospatial pattern cohesion to aggregate patch tokens, which supervises the masked token reconstruction in the student network. This approach makes the task more challenging and effectively enhances the model's ability to learn internal feature distinctions. Our extensive experiments validated PIEViT's efficacy across multiple downstream tasks, including object detection, semantic segmentation, and change detection. The Geospatial Pattern Cohesion (GPC) and Feature Integration Projection (FIP) modules allowed PIEViT to capture complex semantic relationships and distinguish subtle differences within and across images. These results affirm PIEViT's ability to learn rich semantic and local information, demonstrating exceptional generalization and transferability in various remote sensing image interpretation tasks. In the future, we would like to explore more remote sensing data modalities and downstream tasks. For example, we will try to use more types of data during the pre-training process under limited computing power, and we will also try to design self-supervised methods suitable for backbones with larger parameters.

\section*{Acknowledgment}

The authors are graceful to the authors who provided the publicly available datasets.

\ifCLASSOPTIONcaptionsoff
  \newpage
\fi

\end{document}